\documentclass[11pt]{article}

\usepackage[margin=1in]{geometry}
\usepackage{amsmath}
\usepackage{booktabs}
\usepackage{graphicx}
\usepackage{hyperref}
\usepackage{xcolor}
\usepackage[numbers]{natbib}
\usepackage{parskip}
\usepackage{tikz}
\usepackage{float}
\usepackage{mathptmx}
\usepackage{pgfplots}
\pgfplotsset{compat=1.18}
\usetikzlibrary{patterns, decorations.pathreplacing}
\usepgfplotslibrary{groupplots}
\usepackage{soul}
\sethlcolor{yellow!35}
\usepackage{amssymb}
\usepackage{seqsplit}

\makeatletter
\renewenvironment{abstract}{%
  \begin{center}%
    {\bfseries\Large\abstractname}%
  \end{center}%
  \quotation\normalsize
}{%
  \endquotation
}
\makeatother

\title{The Weight of Silence: A Causal Case for Weights Over the Scratchpad in Latent Chess Reasoning}
\author{Ishan S. Kshirsagar \\ \textit{Independent Researcher}}
\date{\today}

\begin{document}
\maketitle

\begin{abstract}
Latent, or \textit{silent}, reasoning lets language models carry out intermediate computation in continuous vector space instead of words, and is widely assumed to function as an internal scratchpad that the model actively consults during inference. Whether that assumption survives reinforcement learning has not been tested directly: existing causal analyses of latent reasoning are confined to math and logic tasks, and compare a model's reliance on its thoughts within a single trained checkpoint, never before and after an RL stage. We train a chess-playing model through a staged latent-reasoning curriculum followed by reinforcement learning, and find that legality climbs monotonically to 61\% (from a 48\% pre-RL baseline) while checkmate confabulation is eliminated entirely. To understand where this gain comes from, we run a six-condition causal intervention suite on the same model before and after reinforcement learning: substituting or adding matched random noise to the latent thought vectors leaves performance unchanged, ablating them (with or without length-matching) causes only mild degradation, and only exact-zero vectors cause collapse. This robustness gap is itself part of the finding: under exact-zero corruption, legality collapses to 1\% pre-RL versus 9\% post-RL, a significant gap that survives correction for testing across the full six-condition battery. Testing the post-RL checkpoint's own side of the battery again at ten times the sample size confirms this pattern holds under substantially more statistical power: removing the thoughts outright and the length-matched removal control both cross into significance as well, in the same severity order the original sample already suggested, while content-preserving substitution and noise remain statistically indistinguishable from undisturbed baseline throughout. Reinforcement learning appears to add robustness to disruption, not reliance on thought content. These results push back against the field's default assumption that latent thoughts function as an actively consulted inference-time scratchpad, and instead indicate that, in this setting, the principal effect of latent reasoning is to shape the model's parameters during training, with much of the downstream improvement encoded in the weights themselves. Alongside this mechanistic result, we demonstrate a working reinforcement-learning gain in chess — a domain outside the math and logic settings where multiple independent groups report the same latent-reasoning-plus-RL recipe failing to improve accuracy over SFT.
\end{abstract}

\newpage
\section{Introduction}
\label{sec:introduction}

Large language models are increasingly trained to carry out part of their reasoning silently, in continuous vector space, rather than only in the words they eventually output. This latent (or silent) reasoning was introduced by Coconut \citep{le2024coconut}, which replaces each intermediate reasoning token with the model's own last hidden state, fed directly as the next input embedding instead of being decoded into a word first. The authors call this a continuous thought, and its central appeal is that it need not collapse into any single vocabulary token: a continuous thought can in principle encode several candidate next steps at
once, letting the model explore more than one line of reasoning in parallel before committing to an answer. That capability is what has made latent reasoning attractive well beyond Coconut itself, and it rests on a specific assumption the field has largely taken as given: that these silent thoughts function as an internal scratchpad the model actively reads from and writes to during inference, the same role an explicit chain-of-thought is assumed to play when a model reasons in words. Whether that assumption survives an RL training stage is not settled, and recent evidence points in different directions depending on the setting studied;
this paper tests it directly in one such setting, comparing a model's reliance on its own latent thoughts before and after reinforcement learning is applied.

To ask this question, we need the specific reinforcement learning recipe being tested. RLVR (reinforcement learning with verifiable rewards) trains a model against an automatically checkable signal, such as whether a move is legal or whether an answer matches a ground truth, rather than a learned reward model. GRPO, introduced by DeepSeekMath \citep{shao2024grpo}, is now the default optimizer for this setting: it scores a group of sampled completions for the same prompt against one another and uses their relative reward as the advantage signal, removing the need for a separately trained critic. However, applying GRPO to a model that reasons partly in latent space has so far not produced the gains it reliably produces for models that reason entirely in words. Ouro, the current state-of-the-art looped language model built around iterative latent computation, reports that RLVR training on top of its supervised checkpoint did not yield significant gains, attributing this to a mix of model saturation and infrastructure limitations in handling the architecture's variable computational depth \citep{zhu2025ouro}. An independent study reaches the same conclusion by a different route: applying GRPO directly to a Coconut-style latent-thinking model produces no improvement over its supervised baseline, because GRPO's update mechanism is built around increasing the probability of intermediate language tokens and has nothing to act on when no such tokens are produced \citep{ozeren2025rl}. Two different architectures, two independent groups, and the same outcome: reinforcement learning, as currently applied, does not appear to help a model that reasons silently.

We find that this null result does not hold universally, though not on the axis those studies measured. We train a chess-playing model through a staged latent-thought curriculum, then apply reinforcement learning on top of it. Accuracy against Stockfish's\footnote{Stockfish is a free, open-source chess engine; we use it throughout this paper as a fixed, deterministic oracle for move legality, move quality, and terminal-position status (checkmate/stalemate), never as a component of the model itself.} own top move stays flat throughout, the same pattern reported in math and logic. The legality climbs monotonically instead, from a 48\% pre-RL baseline to 61\%. A failure mode we term confabulation \citep{farquhar2024semantic} is eliminated entirely: the model no longer confidently declares checkmate on boards where none exists. The central claim of this paper is about where that gain actually lives. We run a battery of causal interventions on the same model before and after reinforcement learning. Neither checkpoint relies on the specific content of its latent thoughts. What changes with
RL is not whether the model uses its thoughts, but how robust it becomes when those thoughts are disrupted, removed, or corrupted outright. In other words, reinforcement learning here does not primarily improve inference-time computation. It reshapes the weights themselves. The improved behavior manifests regardless of what fills the thought positions, indicating that RL's primary effect is to re-weight the base model's parameters rather than teaching the model to actively parse its own latent states during inference.

This finding is based on a six-condition causal battery, run on both checkpoints. Two conditions replace the real latent thoughts with
content that carries no board-specific information: a single fixed vector (substitute) or matched-magnitude noise (noise). Two more remove the thought positions from the sequence, with and without padding it back to its original length (ablate, lenmatch-ablate). A sixth replaces the thoughts with exact zeros (zero). Content-preserving substitutions leave the performance unchanged. Removing the thoughts costs a little. Zeroing them causes performance to collapse.

This pattern holds across both checkpoints: content-invariant, mildly sensitive to removal, catastrophically sensitive to exact zero. That consistency matters: it means that we find no evidence that the model relied on the content of its thoughts, before or after reinforcement learning. What differs is degree, not kind, and every disruptive condition costs the pre-RL checkpoint more than it costs the post-RL checkpoint. Reinforcement learning does not change the purpose of latent thoughts. It changes how much the model needs them to keep functioning.

We test this question in chess. The null results are all from mathematics and logic. Chess offers the same kind of clean, automatically checkable reward those domains rely on: a move is either legal or it is not, and a checkmate claim is either true or it is not. But unlike a single verified numeric answer, chess has no one correct move to imitate, and reasoning about it plays out over a long, adversarial horizon rather than a single step. Reinforcement learning has already been tried in chess, on a model that reasons explicitly in words, and it plateaus there too \citep{hwang2025chess}: the reported ceiling is on move quality, well below expert level, and the authors trace it to a gap in the pretrained model's own understanding of chess that reinforcement learning alone could not close. Our own model shows the same accuracy plateau, evidence for the same underlying limitation. Legality and confabulation sit on a different axis, one that gap does not touch. Reinforcement learning still produces a real, measurable gain there: fewer illegal moves, and no more confabulated checkmates. Running the same causal battery before and after reinforcement learning allows us do more than measure the gain—it lets us locate where it comes from.

\subsection*{Contributions}
\begin{itemize}
  \item \textbf{Behavioral finding.} Legality climbs monotonically to 61\% (from a 48\% pre-RL baseline) through a staged latent-thought curriculum followed by reinforcement learning, surpassing our own explicit chain-of-thought RL baseline (52\%). The move quality itself remains flat throughout: the same accuracy ceiling reported for RL in chess \citep{hwang2025chess} and for RL applied to latent-reasoning models in math and logic \citep{zhu2025ouro, ozeren2025rl}. The gain we report lives on an axis that those settings cannot expose: whether a move is legal, not whether it is optimal.

  \item \textbf{Mechanistic finding.} A six-condition causal battery, run on the same model both before and after reinforcement learning, shows no evidence that legal play depends on the specific content of the model's latent thoughts. What reinforcement learning adds is not reliance on that content, but robustness to disruption -- most clearly established under total signal loss ($1\%$ to $9\%$ retained legality, $p<0.001$), and confirmed by a ten-times-larger replication of the post-RL checkpoint's own side of the battery: removing the thoughts, with or without restoring the original sequence length, costs a small but statistically real amount, while content-preserving substitution and noise remain indistinguishable from undisturbed baseline at any sample size tested.

  \item \textbf{Templated Confabulation.} The model's false-checkmate explanations are not merely incorrect: they are stereotyped, reusing near-identical justification wording across completely unrelated boards (0.68 cross-board text similarity, versus 0.27 for real, distinct move explanations). This is evidence of a fixed, learned template, not board-specific reasoning gone wrong.

  \item \textbf{The Confabulation Cutoff.} Reinforcement learning eliminates checkmate confabulation entirely: 28 to 19 to 0 across the latent-curriculum progression, and separately to 0 in our explicit chain-of-thought RL run as well, two different training recipes, converging on the same outcome independently. The same gated legality reward that drives the behavioral gain above removes this failure mode as a side effect, with no dedicated anti-hallucination signal.

  \item \textbf{Weight-space localization.} The RL-induced weight change is concentrated, not diffuse: 86.2\% of its total energy sits in the MLP projections, led by \texttt{gate\_proj} alone at 56.5\%, against 13.8\% in attention. Within each implicated projection, the change is further concentrated in a small number of directions rather than spread across the LoRA update's full available rank: an exact singular value decomposition of $\Delta W$ at every slot gives a mean effective rank of 2.5 out of a possible 32, and the single largest weight change in the model (layer 19's \texttt{down\_proj}) carries 89\% of its own energy in one direction alone. An independently trained explicit-reasoning RL delta (Rung 1) shows a small but consistently positive structural overlap with this one (whole-adapter cosine similarity $+0.034$), concentrated through the network's first two-thirds and vanishing in its final third. This is evidence that the robustness effect this paper reports is partly a generic consequence of the reward, partly specific to depth, and encoded through a small number of targeted, coherent weight adjustments rather than a diffuse reweighting.
\end{itemize}

\section{Background}

\subsection{Latent reasoning}
\label{sec:background-latent}

Generating a token is usually a two-step process. A model first computes a hidden state at the current position: the output of its final transformer layer, a single real-valued vector the same width as the token embeddings it was built from. That vector is then projected through the language-modeling head and passed through a softmax over the vocabulary, collapsing it into one discrete word. Explicit chain-of-thought reasoning is built entirely out of this second step: every intermediate reasoning step is forced through the same collapse, decoded into a token, written out, and re-read as ordinary text before the model can act on it again. Whatever the hidden state was representing beyond the single word it committed to, a distribution over several plausible next steps, or a genuine ambiguity about which one is right, is discarded at the moment of decoding.

Coconut~\citep{le2024coconut} removes this step for some or all of the reasoning trace. Rather than decoding the hidden state at a reasoning position into a token, Coconut feeds it directly back into the model as the input embedding for the following position, exactly as if it were the embedding of a real word, except no word was ever selected. We follow Coconut's own terminology and call a single instance of this fed-back hidden state a \emph{thought}; \emph{latent thought}, \emph{continuous thought}, and \emph{thought vector} are used interchangeably throughout this paper to refer to the same object. A run of $k$ such positions
replaces $k$ tokens' worth of what would otherwise have been written reasoning, and only the model's final, decoded answer is ever read back as words. The specific number of thought positions used at each stage of this paper's own training curriculum is fixed in \ref{sec:method-training}

The reason to forgo the written trace at all rests on a specific claim about what the vector can carry once it is no longer forced through the softmax. Because a continuous thought never commits to a single vocabulary entry, it can in principle represent more than one candidate continuation at once, several plausible next steps encoded together in the same vector rather than one step chosen, written down, and only afterward found to be wrong. Coconut frames this as an implicit form of breadth-first search folded into a single forward pass. It is this claimed capacity for parallel exploration, not the substitution mechanism itself, that has made latent thoughts widely treated as an active scratchpad the model consults during inference. That assumption is precisely what this paper's causal evidence is built to test.

\subsection{RLVR and GRPO}
\label{sec:background-rlvr}

\emph{Reinforcement learning with verifiable rewards} (RLVR) trains a policy against a reward computed directly from a checkable property of its output, whether a move is legal, whether an answer matches a ground truth, whether a program compiles, rather than a reward model trained to imitate human preferences. Because the signal comes from a fixed procedure rather than a learned or human judge, it cannot itself be flattered by an update that overfits to the judge's own quirks.

GRPO~\citep{shao2024grpo} is the RLVR optimizer used throughout this paper. For a prompt $q$, GRPO samples a group of $G$ completions $\{o_1, \dots, o_G\}$ from the policy prior to the update, $\pi_{\theta_{\mathrm{old}}}$, and scores each with the reward function, producing rewards $r_1, \dots, r_G$. Rather than fit a separate value function to estimate a baseline, as PPO does, GRPO uses the group's own statistics as the baseline: the advantage of completion $i$ is its reward's deviation from the group mean, normalized by the group's standard deviation,
$$
A_i = \frac{r_i - \mathrm{mean}(r_1,\dots,r_G)}{\mathrm{std}(r_1,\dots,r_G)}.
$$
This is the ``group relative'' in the algorithm's name: a completion is rewarded not for its absolute score but for how it compares to other attempts sampled for the exact same prompt, which removes the need for a learned critic entirely. The policy is then updated with a clipped, trust-region objective, in the same spirit as PPO, that pushes each token of each completion toward or away from its current probability in proportion to $A_i$:
$$
\mathcal{J}_{\text{GRPO}}(\theta) = \mathbb{E}\left[\frac{1}{G}\sum_{i=1}^{G}
\frac{1}{|o_i|}\sum_{t=1}^{|o_i|}\min\!\Big(\rho_{i,t}A_i,\
\mathrm{clip}(\rho_{i,t},\,1-\epsilon,\,1+\epsilon)\,A_i\Big)\right]
- \beta\,D_{\mathrm{KL}}\!\left(\pi_\theta \,\|\, \pi_{\mathrm{ref}}\right),
$$
where $\rho_{i,t} = \pi_\theta(o_{i,t} \mid q, o_{i,<t}) \,/\,
\pi_{\theta_{\mathrm{old}}}(o_{i,t} \mid q, o_{i,<t})$ is the token-level
probability ratio between the updated and pre-update policy, $\epsilon$ bounds how far a single update may move it, and $\pi_{\mathrm{ref}}$ is a frozen reference policy the KL term penalizes drift from. The values used at each stage of training are given in \ref{sec:method-training}.

Designing the reward function matters just as much as the optimizer that consumes it. A natural way to reward several sub-goals at once is to \emph{weight} and sum them: some fixed number of points for a legal move, more points for an accurate one, and so on. This is exactly the design most vulnerable to Goodhart's law: when one sub-goal is cheap to satisfy and another is not, the optimizer follows the gradient that is actually easiest to climb, and a policy can drive the weighted sum up by mastering the easy term while the hard term it was meant to encourage stalls or regresses. The alternative used throughout this paper is to \emph{gate} rather than weight: legality is treated as a precondition on receiving any reward at all
rather than as one prize among several, so a response that fails it receives the minimum score outright and every other term goes unevaluated, while a response that clears the gate is then scored on a continuous, dense measure of quality with legality no longer in competition for gradient. This turns ``is the response valid'' into a filter the policy must pass through, not a target it can substitute for the harder goal behind it. The exact gate condition and quality measure used for chess are specified in \ref{sec:method-training}.

\subsection{Mechanistic interpretability and causal intervention}
\label{sec:background-causal}

Two different kinds of evidence can support a claim about what a model's internal representation is doing. Correlational evidence shows that some internal quantity, a neuron's activation, a probe's prediction, a hidden state's projection onto a direction, tracks a property of interest: it rises when the property is present, falls when it is absent. Causal evidence goes further: it shows that changing that internal quantity, and nothing else, changes the model's output in the direction the claim predicts. A representation can correlate with a property the model never
actually uses, information present in the computation but never read from; only an intervention that manipulates the representation directly and observes the downstream effect can distinguish the two. This distinction underlies every claim made about the latent thoughts later in this paper: that a thought vector correlates with the board is not, by itself, evidence that the model's move depends on it.

The standard tool for producing this second kind of evidence is activation
patching, also called causal tracing or interchange intervention. Causal mediation analysis~\citep{vig2020investigating} formalizes this: to test whether some internal component mediates an effect, replace, corrupt, or otherwise hold fixed that component's value during an otherwise normal forward pass, and measure how much of the model's output changes as a result. ROME~\citep{meng2022locating} applies the same logic under the name causal tracing to localize where a model stores a specific factual association, restoring corrupted activations one at a time and reading off which restorations recover the correct output. Ablation is the simplest member of this family: rather than substituting a chosen replacement
value, it removes or zeroes the component entirely and asks what the model can no longer do without it.

None of these interventions is a single, well-defined operation; each requires choosing what to corrupt, what to replace it with, and what to measure, and a systematic comparison across activation-patching methodologies~\citep{zhang2024best} shows that these choices can themselves change the interpretability conclusion drawn from a fixed model and dataset. A single intervention type is accordingly weak evidence on its own: an effect that survives only one specific corruption might reflect that corruption's own side effects rather than the property under study. The causal intervention suite used in this paper (\ref{sec:method-causal-suite}) is built around
this concern directly, applying six distinct interventions of increasing severity to the same latent thoughts rather than trusting the result of any one.

\section{Method}

\subsection{Training ladder}
\label{sec:method-training}
We train four checkpoints in sequence, each building on the last: an SFT baseline, an explicit-reasoning RL checkpoint (Rung 1), a staged latent-reasoning curriculum (Rung 2), and a latent RL checkpoint (Rung 3). Two checkpoints from this ladder carry into every later analysis under the names Table~\ref{tab:headline} uses for them: \textbf{Stage-2}, the pre-RL latent checkpoint, and \textbf{Rung-3}, the post-RL latent checkpoint.

\paragraph{SFT baseline.} We fine-tune Qwen3-14B~\citep{qwen2025qwen3} with LoRA~\citep{hu2021lora} by imitation on 12,002 chess positions. Each position pairs a FEN string, a Stockfish-verified best move, and a written explanation. The explanations are built around deterministic, engine-checked facts (piece identity, captures, checks) rather than facts the generating model had to infer on its own. Every explanation follows a fixed five-section structure, \texttt{[KING SAFETY]}, \texttt{[CHECKS]}, \texttt{[CAPTURES \& TRADES]}, \texttt{[THREATS]}, \texttt{[IMPROVEMENT]}, which later serves as the chunk boundary the latent curriculum dissolves one section at a time. This checkpoint reaches perfect output-format compliance but only 38\% legal moves, with 28 false-checkmate claims out of 100 held-out positions (Table~\ref{tab:headline}). Imitation teaches the vocabulary and structure of chess explanation; it does not teach chess.

\paragraph{Rung 1: explicit reasoning, then GRPO.} We apply GRPO (\ref{sec:background-rlvr}) on top of the SFT checkpoint while reasoning stays in words. The reward gates on legality before it measures quality: an illegal move, an unparseable output, or a false checkmate claim scores $-1.0$ and Stockfish is never called; a correctly identified checkmate scores $+1.0$; any other legal move scores $\exp(-\text{cp\_loss}/120)$, where $\text{cp\_loss} = \max(0,\ \text{best\_eval} - \text{move\_eval})$ is the centipawn gap between the played move and Stockfish's best move, evaluated at depth 12. An earlier version of this reward summed a legality term and a quality term instead of gating one behind the other. Under that design the policy learned to exploit the cheap legality term, and its move quality measurably worsened relative to the SFT checkpoint: the exact Goodharting failure the gated design in \S2.2 exists to prevent. Every Rung 1 number reported in this paper uses the gated version instead (\texttt{GRPO v2} in Table~\ref{tab:headline}): $G=4$ generations per group, learning rate $5\times10^{-6}$, KL coefficient 0.02, 500 steps. It reaches 52\% legal moves with zero false-checkmate claims, the ceiling this paper's latent results are measured against.

\paragraph{Rung 2: staged latent curriculum.} Starting again from the SFT checkpoint, we replace the written reasoning with continuous thoughts (\ref{sec:background-latent}) one section at a time, following Coconut's own recipe: each swallowed section is replaced by two silent thought positions, the optimizer is reset at every stage boundary, and training is pure imitation, with no reward signal, on the same 12,002-position dataset reformatted around this schedule. A stage-$k$ example keeps the first $k$ sections silent, $2k$ thought positions in total, and leaves the remaining sections and the final move in words. The curriculum was designed to run five stages, ending with only the move itself spoken, but training stopped after three. Stage 1 (2 thoughts) reached 46\% legal moves, stage 2 (4 thoughts) reached 48\%, and stage 3 (6 thoughts) dropped to 19\%, tripping the curriculum's own stopping rule, fixed before training began: halt if legality on the frozen harness falls below 20\%. It missed that threshold by a single position. Stages 4 and 5 were never attempted. Stage 2's checkpoint, independently audited to confirm its legal moves were genuine and not an accumulation of lucky false-checkmate declarations, is the \textbf{Stage-2} checkpoint reported throughout this paper and the launch point for Rung 3.

\paragraph{Rung 3: latent GRPO.} We apply GRPO to Stage-2 with the reward unchanged from Rung 1. Only the decoded move is ever scored; the thoughts themselves are not, since there is no ground truth for what a good thought vector looks like. The number of thought positions is fixed at 4, matching Stage-2. The KL reference policy is Stage-2 itself, loaded a second time into a separate, frozen copy, rather than the base model with its adapter disabled: anchoring to raw Qwen3-14B would penalize the policy for retaining anything the latent curriculum had already taught it. Learning rate mattered more here than in Rung 1. A rate of $5\times10^{-6}$ eroded the Stage-2 launchpad, with legality collapsing at the exact steps KL first rose; a rate of $2\times10^{-6}$ moved the policy without destabilizing it. Every Rung 3 result in this paper uses the lower rate, alongside $G=4$ and KL coefficient 0.02, for 150 steps. The resulting checkpoint, \textbf{Rung-3}, reaches 61\% legal moves with zero false-checkmate claims (Table~\ref{tab:headline}), the paper's headline result.

\subsection{Frozen evaluation harness}
\label{sec:method-frozen-eval-harness}
Every number in this paper, from the headline table through the causal battery in \ref{sec:results-add-robustness}, comes from the same evaluation harness: 100 positions fixed once by a single seed (3407) and never changed. Four metrics are recorded for every checkpoint. Format is whether the output has the required tag structure and a move string \texttt{python-chess} can parse, independent of whether that move is legal. Legal is whether the parsed move is legal in the position's FEN. Accuracy is an exact match against Stockfish's own top move at depth 12. A uniformly random legal move already scores close to 10\% here, since a typical position offers roughly ten legal options: accuracy near that floor means no gain in playing strength, not a small one. False mate counts how often the model declares checkmate on a position \texttt{python-chess} confirms is not checkmate.

Decoding is greedy, not sampled, so a checkpoint's score is deterministic given its weights. Latent checkpoints (Stage 1 through 3, and Rung-3) cannot use ordinary text generation, since nothing is decoded at the thought positions. Evaluating them instead uses a separate generation loop: it performs the silent forward-feedback step for the fixed number of thought positions, then resumes ordinary decoding for the rest of the output. The same parsing and scoring code applies afterward, so the four metrics mean the same thing for every checkpoint in the ladder. Latent checkpoints are also run at batch size one. In bf16 precision, a model's output can depend slightly on what else is in its batch, and batch size one removes that variable entirely. The harness itself, meaning the split, the prompts, and the scoring code, was fixed after Rung 1 and reused unchanged for every checkpoint after it, including every causal-intervention condition. No comparison in this paper is confounded by a moving yardstick.

\subsection{Causal intervention suite}
\label{sec:method-causal-suite}
\begin{figure}[H]
\centering
\begin{tikzpicture}[
  every node/.style={font=\small},
  block/.style={draw, minimum width=1.3cm, minimum height=0.8cm, align=center, font=\scriptsize, rounded corners=1pt},
  wide/.style={minimum width=1.8cm},
  promptblock/.style={block, wide, fill=gray!12},
  answerblock/.style={block, wide, fill=gray!12},
  realblock/.style={block, fill=blue!35},
  subblock/.style={block, fill=violet!40},
  noiseblock/.style={block, fill=orange!35, postaction={pattern=north east lines, pattern color=orange!70!black}},
  padblock/.style={block, fill=white, draw=gray, dash pattern=on 2pt off 1.5pt},
  zeroblock/.style={block, fill=white, draw=black, thick},
  rowlabel/.style={font=\bfseries\scriptsize},
  note/.style={font=\scriptsize\itshape, gray}
]

\def\xprompt{0}
\def\xt{1.9}
\def\dt{1.5}
\def\xanswer{8.3}
\def\xlabel{-1.2}

\node[rowlabel, anchor=east] at (\xlabel,0)    {Baseline};
\node[rowlabel, anchor=east] at (\xlabel,-1.3) {Substitute};
\node[rowlabel, anchor=east] at (\xlabel,-2.6) {Noise};
\node[rowlabel, anchor=east] at (\xlabel,-3.9) {Ablate};
\node[rowlabel, anchor=east] at (\xlabel,-5.2) {Lenmatch-Ablate};
\node[rowlabel, anchor=east] at (\xlabel,-6.5) {Zero};

\node[promptblock] at (\xprompt,0) {prompt};
\node[realblock] at (\xt,0) {$t_1$};
\node[realblock] at (\xt+\dt,0) {$t_2$};
\node[realblock] at (\xt+2*\dt,0) {$t_3$};
\node[realblock] at (\xt+3*\dt,0) {$t_4$};
\node[answerblock] at (\xanswer,0) {move};

\node[promptblock] at (\xprompt,-1.3) {prompt};
\node[subblock] at (\xt,-1.3) {$T^*$};
\node[subblock] at (\xt+\dt,-1.3) {$T^*$};
\node[subblock] at (\xt+2*\dt,-1.3) {$T^*$};
\node[subblock] at (\xt+3*\dt,-1.3) {$T^*$};
\node[answerblock] at (\xanswer,-1.3) {move};

\node[promptblock] at (\xprompt,-2.6) {prompt};
\node[noiseblock] at (\xt,-2.6) {$n_1$};
\node[noiseblock] at (\xt+\dt,-2.6) {$n_2$};
\node[noiseblock] at (\xt+2*\dt,-2.6) {$n_3$};
\node[noiseblock] at (\xt+3*\dt,-2.6) {$n_4$};
\node[answerblock] at (\xanswer,-2.6) {move};

\node[promptblock] at (\xprompt,-3.9) {prompt};
\node[answerblock] at (\xt+1.9,-3.9) {move};
\draw[gray, <->, dashed] (0.95,-3.9) -- (2.85,-3.9);
\node[fill=white, inner sep=1pt, font=\tiny\itshape, text=gray] at (1.9,-3.9) {4 slots removed};

\node[promptblock] at (\xprompt,-5.2) {prompt};
\node[padblock] at (\xt,-5.2) {PAD};
\node[padblock] at (\xt+\dt,-5.2) {PAD};
\node[padblock] at (\xt+2*\dt,-5.2) {PAD};
\node[padblock] at (\xt+3*\dt,-5.2) {PAD};
\node[answerblock] at (\xanswer,-5.2) {move};

\node[promptblock] at (\xprompt,-6.5) {prompt};
\node[zeroblock] at (\xt,-6.5) {$\vec{0}$};
\node[zeroblock] at (\xt+\dt,-6.5) {$\vec{0}$};
\node[zeroblock] at (\xt+2*\dt,-6.5) {$\vec{0}$};
\node[zeroblock] at (\xt+3*\dt,-6.5) {$\vec{0}$};
\node[answerblock] at (\xanswer,-6.5) {move};

\draw[decorate, decoration={brace, amplitude=4pt}, thick] (\xt-0.4, 0.55) -- (\xt+3*\dt+0.4, 0.55);
\node[above, font=\scriptsize] at (\xt+1.5*\dt, 0.75) {4 latent thought-vector slots};

\end{tikzpicture}
\caption{The six-condition causal intervention suite. Each row shows the same underlying sequence: a chess board prompt, four latent thought-vector positions, and the model's move, with a different intervention applied to the four thought slots (highlighted).}
\label{fig:causal_setup}
\end{figure}
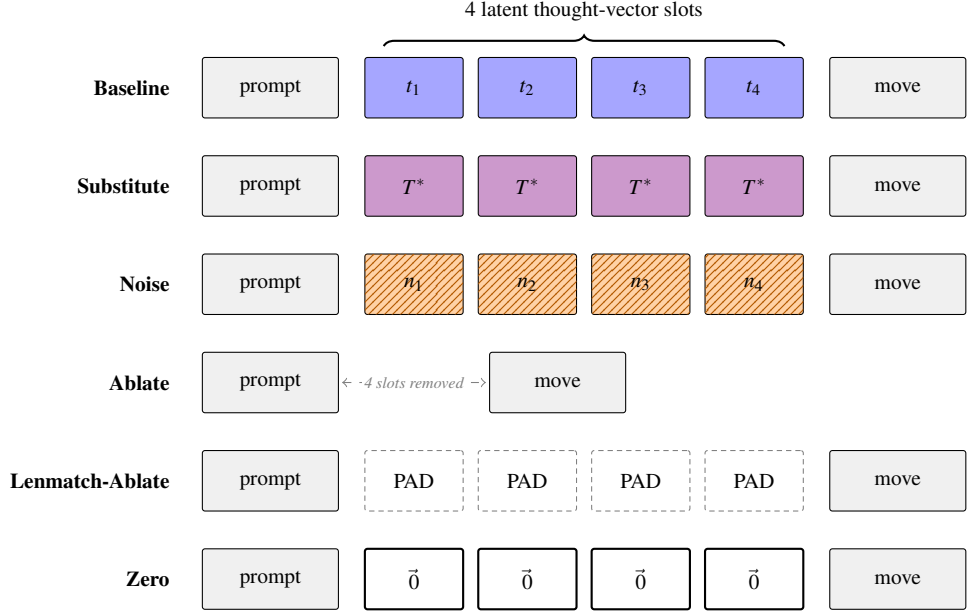

A single kind of intervention is weak evidence on its own: an effect that only shows up under one specific corruption might reflect that corruption's own side effects, not the property being tested (\ref{sec:background-causal}). We apply six conditions to the four thought positions instead, ordered from a no-op control to the most severe manipulation we test, matching the increasing-severity design described in \ref{sec:background-causal}, and we run the identical suite on Stage-2 and Rung-3 alike (Figure~\ref{fig:causal_setup}), so any pattern in the results can be attributed to reinforcement learning rather than to which corruption happened to be used.

\textbf{Baseline} runs the model exactly as trained: the four thought positions hold the real, board-conditioned hidden states produced by the model's own forward pass.

\textbf{Substitute} replaces all four positions with a single fixed vector $T^*$, computed once by averaging real thought vectors from 30 reference boards held out of the evaluation set. The same $T^*$ fills every position of every evaluated board, so nothing about it varies with the board in front of the model or with which of the four positions it fills. This tests whether the thoughts' specific content matters at all, while leaving their scale and the sequence's length untouched.

\textbf{Noise} also replaces all four positions, but with an independent random vector at each one: a random direction, rescaled to match $T^*$'s own norm. Where Substitute is content-free but fixed, Noise is content-free and different at every position and every board, isolating whether even a consistent direction matters, or only the right scale does.

\textbf{Ablate} removes the four thought positions from the sequence outright. No forward pass is run for them, and the sequence is four positions shorter than baseline. This tests presence against absence rather than content.

\textbf{Lenmatch-Ablate} keeps the same removal but restores the original sequence length, filling the four positions with the model's own pad-token embedding, attention unmasked, so the model actually attends to them rather than ignoring them as it does throughout training. It asks whether any gap from plain ablation is a raw length artifact rather than a consequence of removing the thoughts themselves.

\textbf{Zero} keeps all four positions in the sequence but fills them with the exact-zero vector, a magnitude the model never encounters at any other point in training or inference. It tests tolerance for an input shape the model has never seen, not content.

Together the six conditions separate three questions: whether thought content matters (Substitute, Noise), whether the positions need to exist at all (Ablate, Lenmatch-Ablate), and whether the model can tolerate an input it has never encountered (Zero). Results for all six, on both checkpoints, are in Table~\ref{tab:causal} and \ref{sec:results-add-robustness}.

\subsection{Experiment 2: Gumbel reparameterization}
\label{sec:method-gumbel}

Rung 3's thoughts are deterministic. At each thought position, the model's own last-layer hidden state is fed back directly, computed once and never resampled. GRPO's gradient therefore only ever reaches the words spoken after the thoughts, never the thoughts themselves. We adapt this experiment from the Gumbel-reparameterized recipe SofT-GRPO~\citep{zheng2025soft} uses to give RL a gradient into soft-thinking tokens. It gives reinforcement learning an actual path into thought content, testing whether a working gradient there lets RL discover something the fully deterministic recipe could not, rather than converging to the same board-blind pattern regardless.

At each of the four thought positions, we take the model's own next-token logits and mask them to the top 20 candidates. We scale by $1/\tau$ with $\tau=4$, add Gumbel noise, and pass the result through a softmax, the Gumbel-Softmax reparameterization~\citep{jang2016gumbel}. The forward pass then hard-argmaxes to a single real vocabulary token and feeds that token's own embedding back in, so the model always conditions on a genuine, previously-seen embedding, never a blurred mixture. The backward pass flows through the soft distribution behind the pick instead, the standard straight-through trick. The reference policy for the KL term never uses this mechanism: it stays on Rung 3's original deterministic path throughout, so the KL leash still anchors to genuine Stage-2 behavior. KL is scored on spoken tokens only, exactly as in Rung 3, since a KL term between two different thought mechanisms, discrete Gumbel picks against continuous deterministic vectors, is not well defined.

We reached $\tau=4$ with top-20 masking after two failed settings. $\tau=1$ with no masking collapsed to near-zero variation: the model's own confidence gaps are large enough that its top choice wins under fair sampling almost every time. $\tau=8$ with no masking produced incoherent output instead, since softening the entire roughly 150,000-word vocabulary lets noise drawn from that whole space overwhelm any real confidence gap. Masking to the top 20 candidates before adding noise keeps the comparison meaningful while still producing real variation.

We launch from Stage-2 and train for 150 steps, matching Rung 3's budget exactly so the mechanism is the only variable between the two runs. Switching to sampled thought tokens costs some legality immediately: the run starts at 44\% legal, below Stage-2's own 48\%, since Stage-2 was trained expecting raw hidden-state feedback rather than sampled-token embeddings. Legal-move rate then rises to 50\% by step 150 (\ref{sec:results-gumbel}).

\section{Results}

\subsection{Headline results}
\label{sec:results-headline}

Table~\ref{tab:headline} reports all four checkpoints from the training progression in \ref{sec:method-training}, scored on the frozen 100-position harness in \ref{sec:method-frozen-eval-harness}. Legal-move rate climbs from 38.0\% (SFT, imitation only) through 52.0\% (GRPO v2, explicit reasoning + RL) and 48.0\% (Stage-2, latent reasoning with no RL yet) to 61.0\% (Rung-3, latent curriculum + RL): only the latent-curriculum-then-RL recipe reaches the highest legal-move rate of the four. Both RL-trained checkpoints eliminate false-checkmate claims entirely, 28 to 0 for GRPO v2 and 19 to 0 for Rung-3, but Stage-2 alone only partially reduces them, underscoring that reinforcement learning, not the latent curriculum by itself, drives both gains. Rung-3's 61\% is the endpoint of its own RL run, not a single jump: legality reaches 56\% at step 100, 60\% at step 125, and 61\% at step 150, where training stopped. Accuracy, an exact match against Stockfish's own top move, stays flat at 9--10\% across every checkpoint: no stage of training meaningfully changes how often the model finds the objectively best move. The gain reported throughout this paper is a legality and confabulation story, not a claim that the model plays stronger chess. Figure~\ref{fig:headline_trend} plots the resulting trend across the full progression, alongside the parallel collapse in false-checkmate claims.

\begin{table}[h]
\centering
\begin{tabular}{lcccc}
\toprule
Model & Format \% & Legal \% & Accuracy \% & False Mate \\
\midrule
SFT (imitation)            & 100.0 & 38.0 & 9.0  & 28 \\
GRPO v2 (explicit + RL)    & 100.0 & 52.0 & 10.0 & 0  \\
Stage-2 (latent, no RL)    & 100.0 & 48.0 & 10.0 & 19 \\
\textbf{Rung-3 (latent + RL)} & 100.0 & \textbf{61.0} & 9.0 & \textbf{0} \\
\bottomrule
\end{tabular}
\caption{Headline results across the four-checkpoint training progression (\ref{sec:method-training}): SFT (imitation baseline), GRPO v2 (explicit chain-of-thought reasoning + RL), Stage-2 (latent-thought curriculum, pre-RL), and Rung-3 (latent curriculum + RL, this paper's headline checkpoint). Format is compliance with the required output structure; Legal is legality of the parsed move in the position's FEN; Accuracy is exact match against Stockfish's top move at depth 12; False Mate counts confident checkmate declarations on positions that are not checkmate. Metric definitions in full in \ref{sec:method-frozen-eval-harness}.}
\label{tab:headline}
\end{table}

\pgfplotsset{
  causalbarstyle/.style={
    ybar,
    bar width=18pt,
    width=0.5\textwidth,
    height=6cm,
    xtick=data,
    x tick label style={rotate=20, anchor=east, font=\small},
    ymajorgrids=true,
    grid style={gray!25},
    axis line style={gray!60},
    tick style={gray!60},
    enlarge x limits=0.25,
    nodes near coords,
    every node near coord/.style={font=\small},
  }
}

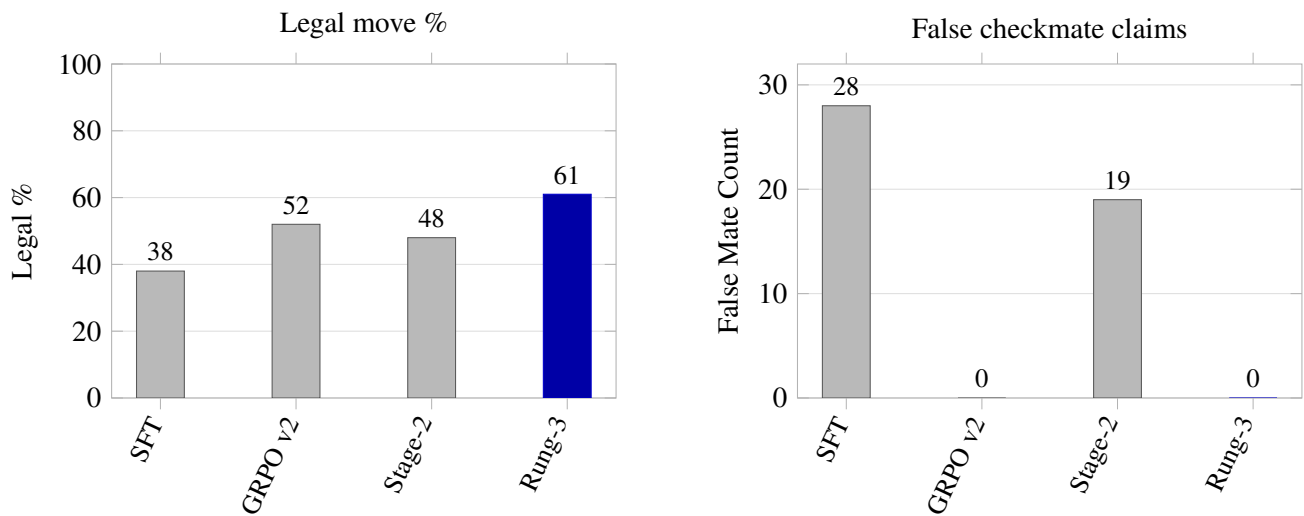
\begin{figure}[H]
\centering
\begin{tikzpicture}
\begin{groupplot}[
  group style={group size=2 by 1, horizontal sep=2.4cm},
]

\nextgroupplot[
  causalbarstyle,
  bar shift=0pt,
  enlarge x limits=0.12,
  x tick label style={rotate=45, anchor=east, font=\small},
  xtick={0,1,2,3},
  xticklabels={SFT,GRPO v2,Stage-2,Rung-3},
  ymin=0, ymax=100,
  ytick={0,20,40,60,80,100},
  ylabel={Legal \%},
  title={Legal move \%},
]
\addplot[fill=gray!55, draw=gray!70!black] coordinates {(0,38)};
\addplot[fill=gray!55, draw=gray!70!black] coordinates {(1,52)};
\addplot[fill=gray!55, draw=gray!70!black] coordinates {(2,48)};
\addplot[fill=blue!65!black, draw=blue!80!black] coordinates {(3,61)};

\nextgroupplot[
  causalbarstyle,
  bar shift=0pt,
  enlarge x limits=0.12,
  x tick label style={rotate=45, anchor=east, font=\small},
  xtick={0,1,2,3},
  xticklabels={SFT,GRPO v2,Stage-2,Rung-3},
  ymin=0, ymax=32,
  ytick={0,10,20,30},
  ylabel={False Mate Count},
  title={False checkmate claims},
]
\addplot[fill=gray!55, draw=gray!70!black] coordinates {(0,28)};
\addplot[fill=gray!55, draw=gray!70!black] coordinates {(1,0)};
\addplot[fill=gray!55, draw=gray!70!black] coordinates {(2,19)};
\addplot[fill=blue!65!black, draw=blue!80!black] coordinates {(3,0)};

\end{groupplot}
\end{tikzpicture}
\caption{Legal-move percentage (left) and false checkmate count (right) across the four models in Table~\ref{tab:headline}, same order. Navy marks Rung-3, the paper's headline checkpoint (latent curriculum + RL); gray marks the other three. Both explicit-CoT RL (GRPO v2) and latent+RL (Rung-3) eliminate false checkmate claims entirely, but only the latent-curriculum-then-RL recipe also reaches the highest legal-move rate. Stage-2 -- latent reasoning with no RL yet -- trails both RL-trained checkpoints on legality and only partially reduces false mates, underscoring that RL, not the latent curriculum alone, drives both gains.}
\label{fig:headline_trend}
\end{figure}

\subsection{Templated Confabulation}
\label{sec:results-tempconfab}

\begin{figure}[H]
\centering
\begin{minipage}[t]{0.48\textwidth}
\small\raggedright
\fbox{\begin{minipage}{0.94\textwidth}
\raggedright
\textbf{Board A} \\
\texttt{\tiny r3r1k1/pp2R1Bp/3bB1p1/2p1n3/4p1P1/7P/PPP1K3/7R b - - 0 23}

\smallskip
[CAPTURES \& TRADES]: \hl{Material is irrelevant} as \hl{the game is over due to the checkmate condition.}

\smallskip
{[THREATS]:} The Rook on e7 \hl{delivers a perpetual check, trapping the Black King with no escape or defense.}

\smallskip
{[IMPROVEMENT]:} \hl{No further moves are possible for Black}; all potential responses would result in an immediate loss.

\smallskip
\hl{Conclusion: This position is CHECKMATE. White wins.}
\end{minipage}}
\end{minipage}
\hfill
\begin{minipage}[t]{0.48\textwidth}
\small\raggedright
\fbox{\begin{minipage}{0.94\textwidth}
\raggedright
\textbf{Board B} \\
\texttt{\tiny 8/7R/1p3p2/4r3/1P6/Pk1K2P1/2r2P1P/3R4 b - - 2 39}

\smallskip
[CAPTURES \& TRADES]: \hl{Material is irrelevant} because \hl{the game is over due to the checkmate condition.}

\smallskip
{[THREATS]:} The Rook on h7 \hl{delivers a perpetual check, trapping the Black King with no escape or defense.}

\smallskip
{[IMPROVEMENT]:} \hl{No further moves are possible for Black} as it has 0 legal moves and is in check.

\smallskip
\hl{Conclusion: This position is CHECKMATE. White wins.}
\end{minipage}}
\end{minipage}
\caption{Two false-checkmate completions from unrelated boards, sampled from the stage-2 (pre-RL) causal battery -- Board A from the \texttt{baseline} condition, Board B from the \texttt{substitute} condition. Highlighted spans mark wording identical word-for-word across both completions; unhighlighted text marks the only board-specific insertion (the threatening piece and its square). The four-section structure, the ``Material is irrelevant'' opener, the ``delivers a perpetual check... no escape or defense'' clause, and the closing ``Conclusion'' line all reappear verbatim. Board A is itself not actually checkmate -- Black had a legal escape (e5f7) that this and three other non-zero causal conditions all failed to find, each producing an equally confident, equally wrong justification.}
\label{fig:templated_confabulation}
\end{figure}

Figure~\ref{fig:templated_confabulation} shows two such completions side by side. Beyond the four-section structure every completion in this dataset shares, the wording itself repeats almost verbatim between them: the same ``Material is irrelevant'' opener, the same ``delivers a perpetual check... no escape or defense'' clause, the same closing ``Conclusion'' line. The only content that changes between the two boards is the threatening piece and its square, inserted into an otherwise fixed sentence.

To check whether this pair was representative rather than hand-picked, we measured pairwise text similarity across every false-checkmate completion produced during the stage-2 causal battery, 96 completions in total, summed across the six conditions in Table~\ref{tab:causal}. Rung-3 offers none to compare, since its false-mate count is zero in every condition. False-checkmate explanations average 0.68 similarity to one another. Real, distinct move explanations sampled across different boards average only 0.27, despite sharing the identical five-section template. Both groups are built on the same fixed scaffold, so the gap between them is content, not format. An earlier pass through this analysis paired same-board completions from different causal conditions against each other by mistake, artificially inflating both figures; excluding those pairs and keeping only genuinely cross-board comparisons barely moved either number, evidence the pattern is real rather than a counting artifact.

We call this pattern \textbf{Templated Confabulation}: rather than reasoning freshly about each board, the model falls back on a small set of memorized justification templates, most visibly at the exact moment it is about to be wrong. An error that tracked the board, a genuine but mistaken read of the position, would vary with the board. This one does not. That is evidence of a fixed, learned script triggered by surface features of the position, not board-specific reasoning gone wrong.

\subsection{The Confabulation Cutoff}

Table~\ref{tab:headline} shows checkmate confabulation dropping from 28 (SFT) to 19 (Stage-2, latent reasoning with no RL yet) to 0 in both reinforcement-learned checkpoints, GRPO v2 and Rung-3 alike. The two zeros are independent results, not one: one comes from RL applied to explicit chain-of-thought reasoning, the other from RL applied on top of the latent-thought curriculum. Two different training recipes converge on the same outcome.

No term in the reward function (\ref{sec:background-rlvr}, \ref{sec:method-training}) targets confabulation directly. A false checkmate claim is scored exactly like an illegal move or an unparseable output: $-1.0$, with Stockfish never consulted. Confabulation is not a separate failure mode the reward has to learn to recognize on its own terms; it falls under the same legality gate that catches every other invalid response. Eliminating it is a side effect of optimizing against that gate, not a target the training process was built to pursue directly.

The two drops in Table~\ref{tab:headline} have different causes. The fall from 28 to 19 happens before any reward signal is applied at all: replacing written explanations with silent thought positions during the latent curriculum (\ref{sec:method-training}), pure imitation, no RL, roughly halves the behavior on its own. The fall from 19 to 0 is what reinforcement learning contributes, and only the gated legality reward finishes the job, reaching zero twice, independently, under two different reasoning formats.

One caveat applies to every number in this subsection. The frozen evaluation harness (\ref{sec:method-frozen-eval-harness}) contains no genuine checkmate position among its 100 boards, confirmed by a separate audit of every legal move logged at every stage of training. Every checkmate declaration recorded anywhere in this paper is therefore necessarily false, and every reduction in the false-mate count reflects the model learning to withhold an incorrect claim, not learning to correctly identify a real one. No checkpoint in this paper, including Rung-3, has ever been given the opportunity to call a real checkmate, so we have no evidence either way on whether it could.

One number in this section is itself worth qualifying with the same caution just applied to it. Every reduction reported above rests on a 100-position sample; a large-sample replication of Rung-3's own baseline condition at $n=1000$ (Appendix~\ref{app:n1000}) finds 5 false-checkmate claims rather than 0, a residual rate of roughly 0.5\%. This does not revise the headline harness figure in Table~\ref{tab:headline}, which remains exactly 0 at $n=100$ and was not rerun at this scale; it does mean that the 0-of-100 result from the causal battery's own baseline condition (Table~\ref{tab:causal}) reflects a very low true rate rather than a provably exact one, a distinction 100 positions alone cannot resolve.

\subsection{RL adds robustness, not content-reliance}
\label{sec:results-add-robustness}

Rather than run the six-condition battery from \ref{sec:method-causal-suite} once and infer what reinforcement learning changed from elimination logic alone, content doesn't seem to matter, so it must be the weights, we run the identical battery on Stage-2 and Rung-3: the same six manipulations, applied to the same four thought positions, scored on the same frozen 100-position harness (\ref{sec:method-frozen-eval-harness}), on a checkpoint before reinforcement learning and on that checkpoint's own descendant after it. Nothing about the intervention or the evaluation changes between the two runs. Any difference between a Stage-2 cell and its matched Rung-3 cell in Table~\ref{tab:causal} can only come from one place: what reinforcement learning did to the weights in between.

\begin{table}[h]
\centering
\begin{tabular}{lcccc}
\toprule
Condition & Stage-2 Legal \% & Rung-3 Legal \% & Stage-2 FMate & Rung-3 FMate \\
\midrule
Baseline          & 48.0 & 58.0 & 19 & 0 \\
Substitute        & 48.0 & 58.0 & 20 & 0 \\
Noise             & 48.0 & 60.0 & 17 & 0 \\
Ablate            & 44.0 & 57.0 & 22 & 0 \\
Lenmatch-Ablate   & 39.0 & 53.0 & 18 & 0 \\
Zero              & 1.0  & 9.0  & 0  & 0 \\
\bottomrule
\end{tabular}

\smallskip
{\small\itshape Note: Rung-3's baseline legal rate here (58\%) differs from the headline figure in Table~\ref{tab:headline} (61\%). Both come from the same adapter; the gap is normal run-to-run variation from NF4 kernel dequantization order under bf16, not a discrepancy in the underlying result. Every comparison within this table uses its own internally consistent baseline (58\%), not the headline number. A large-sample replication of this same baseline condition at $n=1000$ (Appendix~\ref{app:n1000}) gives 56.7\%, in the same range as both figures here and consistent with ordinary evaluation-run variation rather than any systematic drift.}

\caption{Six-condition causal battery (\ref{sec:method-causal-suite}) on the frozen 100-position harness (\ref{sec:method-frozen-eval-harness}), run separately on Stage-2 (pre-RL) and Rung-3 (post-RL). Legal is legality of the parsed move; FMate counts confabulated checkmate declarations, both defined as in Table~\ref{tab:headline}.}
\label{tab:causal}
\end{table}

\begin{figure}[H]
\centering
\begin{tikzpicture}
\begin{axis}[
  ybar,
  bar width=13pt,
  width=0.95\textwidth,
  height=6.5cm,
  ymin=0, ymax=100,
  ylabel={Legal \%},
  symbolic x coords={Baseline,Substitute,Noise,Ablate,Lenmatch-Ablate,Zero},
  xtick=data,
  x tick label style={rotate=20, anchor=east, font=\small},
  ytick={0,20,40,60,80,100},
  ymajorgrids=true,
  grid style={gray!25},
  axis line style={gray!60},
  tick style={gray!60},
  enlarge x limits=0.12,
  legend style={at={(0.5,1.15)}, anchor=south, legend columns=-1, draw=none, font=\small},
  nodes near coords,
  every node near coord/.append style={font=\tiny, /pgf/number format/.cd, fixed, precision=0},
]
\addplot[fill=gray!55, draw=gray!70!black] coordinates {
  (Baseline,48) (Substitute,48) (Noise,48) (Ablate,44) (Lenmatch-Ablate,39) (Zero,1)
};
\addplot[fill=blue!65!black, draw=blue!80!black] coordinates {
  (Baseline,58) (Substitute,58) (Noise,60) (Ablate,57) (Lenmatch-Ablate,53) (Zero,9)
};
\legend{Stage-2 (pre-RL), Rung-3 (post-RL)}
\end{axis}
\end{tikzpicture}
\caption{Legal-move percentage across all six causal conditions, plotted from Table~\ref{tab:causal}. Gray bars: Stage-2 (pre-RL). Navy bars: Rung-3 (post-RL).}
\label{fig:causal_battery_chart}
\end{figure}
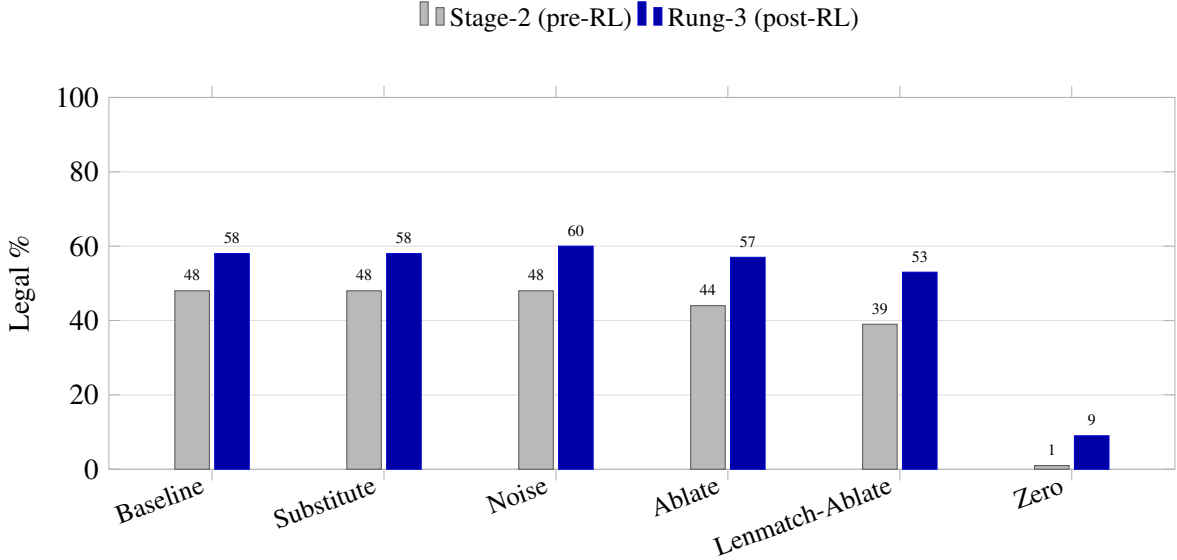

The first three conditions in Figure~\ref{fig:causal_battery_chart}, Baseline, Substitute, and Noise, sit at nearly the same height within each checkpoint: replacing the real thought content with a fixed vector or with matched random noise costs almost nothing, on Stage-2 as much as on Rung-3. This rules out one story before it can be told. Reinforcement learning did not teach the model to stop relying on its thoughts, because Stage-2 already shows the same content-invariance before any RL touched it. The model never leaned on the specific content of its thoughts in the first place.

The real finding sits across the full five-condition retention-gap analysis, and its
strength differs by condition. The point estimates trace a monotonic staircase in
exactly the severity order the battery was designed around: Substitute $+0.000$,
Noise $+0.034$, Ablate $+0.066$, Lenmatch-Ablate $+0.101$, Zero $+0.134$ -- each
condition costing Rung-3 proportionally less of its own baseline competence than it
costs Stage-2, by a growing margin as the disruption grows more severe. Only Zero
clears significance: both checkpoints collapse under it (McNemar $p<0.001$ for
each), and a bootstrap 95\% confidence interval on the retention-ratio gap excludes
zero entirely ($+0.134$, CI $[0.050, 0.231]$), meaning Rung-3 retains a
significantly larger share of its own competence under total collapse. This result
survives a Bonferroni correction for testing all five conditions as a family as
well (CI $[0.031, 0.263]$ at a 99\%-per-test threshold), so the finding is not an
artifact of reporting the one test out of five that happened to clear an
uncorrected bar. Lenmatch-Ablate shows a narrower asymmetry: the intervention
significantly hurts Stage-2 on its own (McNemar $p=0.023$, versus $p=0.27$ for
Rung-3), though the between-checkpoint retention gap does not itself clear
significance under either correction (CI $[-0.052, 0.256]$). Substitute and Noise,
the two content-preserving conditions, sit at or near a zero gap ($+0.000$ and
$+0.034$), consistent with this section's earlier point that neither checkpoint
relies on thought content in the first place. Ablate falls between the two
extremes in the predicted direction but likewise does not reach significance at
$n=100$. Full test statistics for all five conditions, uncorrected and
Bonferroni-corrected, are in Appendix~\ref{app:significance}. A large-sample replication of Rung-3's own side of this battery at $n=1000$, testing whether this $n=100$ ceiling on significance reflects a genuinely absent effect or simply limited power, is in Appendix~\ref{app:n1000}.

Aggregate legal rate hides some of this. Pairing each completion against its counterpart on the exact move played, not just whether it was legal, shows baseline-vs-noise agreement at 91\% on Rung-3 but only 82\% on Stage-2. The two checkpoints reach nearly the same pass rate under noise, but Stage-2 gets there with more individual answers flipping from one attempt to the next. Reinforcement learning did not only preserve the pass rate under disruption, it locked in which move gets played, at the level of individual positions, not only in aggregate.

\subsection{J-lens: a fixed, board-invariant scaffold}
\label{sec:results-jlens}

Section \ref{sec:results-add-robustness} shows behaviorally that legal play does not depend on the content of the latent thoughts: replacing them costs almost nothing, on either checkpoint. That result is silent on a separate question: how much board-specific content is sitting in those thought vectors in the first place, whether or not the model ends up using it. This subsection answers that question directly.

We use the Jacobian lens (J-lens) \citep{gurnee2026jlens}, an interpretability technique that reads a hidden state at a given layer and position and reports which vocabulary tokens it is currently disposed to produce, ahead of the model actually saying them. Where the older logit lens \citep{nostalgebraist2020logitlens} applies the model's own unembedding directly to an intermediate hidden state, J-lens first transports that hidden state into the final-layer basis using a fitted Jacobian, then unembeds it, a transport step the plain logit lens skips entirely. We fit the lens directly on Rung-3 in its true form, the 4-bit quantized base model with the trained LoRA adapter attached live, rather than on a merged, full-precision copy. Merging first would mean analyzing a model that never existed during training, since the adapter was trained specifically to correct for 4-bit quantization error.

Before pointing the lens at anything latent, we calibrated it on ordinary written chess reasoning, sweeping across read-depth to find where it reliably recovers real content, since the right layer is model- and scale-dependent rather than something to assume from prior work (\ref{sec:background-causal}). Layers 26 through 29 turned out to be the useful band: deep enough that the model has already settled on which piece is relevant, shallow enough that the plain logit lens has not yet caught up. At layer 28, J-lens recovers the upcoming piece\footnote{Restricted to King, Queen, Knight, Bishop, and Pawn. Rook is multi-token under this tokenizer and unreadable by either lens; squares and full SAN moves fragment even further and were not attempted.} roughly 94\% of the time, against roughly 0\% for the plain logit lens at the same layer, a gap of about 90 points. By layers 34 through 38 both methods converge near 100\%, since by that depth the piece has effectively already been produced and there is nothing left to anticipate.

Reading the calibrated band at the four thought positions themselves returned something close to noise: largely incoherent output, and suspiciously similar regardless of which board produced it. To check that directly, independent of how well the lens happens to decode it, we measured plain cosine similarity between the raw thought vectors across genuinely different boards. The result: approximately 0.99 similarity, board after board. Slot-to-slot within a single board is far less similar, 0.31 to 0.64, so the four positions are not one vector repeated four times, they form a distinct four-step pattern. But that pattern barely changes with the board in front of the model. It is a fixed, learned scaffold, not a board-conditioned representation.

The calibration step is what makes this null result trustworthy rather than merely inconclusive. A lens that already performs badly on ordinary text would make an empty reading of the thoughts uninformative, indistinguishable from the tool simply not working. J-lens does not have that problem here: it recovers real content at roughly 94\% on text from the same domain, at the same depth, on the same model. Its failure to find comparable content in the latent thoughts is therefore evidence about what is and is not there, not evidence about the lens. Combined with \ref{sec:results-add-robustness}, two independent methods now point the same way: a behavioral one, ablating or substituting the thoughts barely moves legal-move rate, and a mechanistic one, there is little board-specific signal in the thoughts to begin with. Neither the model's behavior nor its internal representations show it relying on the specific content of its own silent reasoning.

One further check is worth running before treating this collapse as evidence about
training rather than architecture. We repeated the diagnostic across the full
four-checkpoint progression -- the raw base model, the SFT checkpoint, Stage-2, and
Rung-3 -- computing the same cross-board cosine similarity for each, now averaged over
the full 100-position harness rather than the four boards used above (Appendix~\ref{app:basemodel}
reports every value; Rung-3 reproduces the $\approx 0.99$ figure above when measured on
the original four boards, confirming the two runs are consistent). The base model
already sits at 0.874, and SFT drops slightly to 0.812. Neither figure should be read as
evidence of an architectural bias toward board-invariant representations on its own,
since feeding a model's own hidden state back in as an input embedding is an operation
neither checkpoint was ever trained to perform, and both numbers may partly reflect that
unfamiliarity rather than a genuine representational tendency. The comparison that is
not subject to this caveat is Stage-2 against Rung-3, since both were trained directly
on this mechanism: cosine similarity is 0.992 at Stage-2 and 0.992 at Rung-3, identical
to three decimal places. The collapse into a fixed, board-invariant scaffold is
therefore complete by the end of the latent curriculum, before reinforcement learning
is ever applied. Reinforcement learning's contribution is not to this representation;
it is the separate robustness effect established causally in \ref{sec:results-add-robustness}. Imitation training
on the latent curriculum builds the fixed scaffold; reinforcement learning determines
how much the surrounding network needs that scaffold intact in order to keep
functioning. Before concluding, we test whether a working gradient into thought content would change this picture.

\subsection{Experiment 2: Gumbel result}
\label{sec:results-gumbel}

As \ref{sec:method-gumbel} describes, this run starts at 44\% legal, below Stage-2's own 48\%, purely from switching the thought mechanism: Stage-2 was trained expecting raw hidden-state feedback, and the discrete, Gumbel-sampled token embeddings used here are a genuine distribution shift at step 0. From that lower starting point, legal-move rate climbs to 46\% at step 50, 47\% at step 100, and 50\% at step 150, the same budget used for Rung-3's own run.

Two readings of that endpoint matter, and they point in different directions. Read against the other checkpoints directly, 50\% sits barely above Stage-2's 48\% and well below both GRPO v2 (52\%) and Rung-3 (61\%): giving reinforcement learning a working gradient into thought content did not let it discover anything Rung-3's fully deterministic recipe missed. Read against its own starting point instead, the climb is 6 points in 150 steps, roughly half of Rung-3's 13-point climb over the identical budget, even after accounting for the mechanism-swap cost at step 0. Either way, the result lands in Stage-2's territory, not Rung-3's.

Confabulation tells a less clean story here than legality does. The false-mate count falls from 26 at step 0 to 18 at step 50 and 15 at step 100, where it plateaus: step 150 still reports 15. Unlike GRPO v2 and Rung-3, both of which reach zero, this run does not eliminate confabulation within the matched budget. Accuracy stays flat at 8--9\% throughout, consistent with every other checkpoint in this paper.

This is a second independent line of evidence that points in the same direction as \ref{sec:results-add-robustness}. Giving reinforcement learning a working gradient into thought content did not produce a better policy than Rung-3's fully deterministic recipe, whose gradient never reaches the thoughts at all; it produced a worse one, converging toward roughly Stage-2's own territory instead. One caveat applies honestly rather than being smoothed over: the straight-through Gumbel estimator is a known-biased gradient approximation, so this result cannot cleanly separate two different explanations, that thought content genuinely does not help, or that this particular optimization path is structurally harder to train regardless of content. This paper reports the result at the matched 150-step budget it was designed to compare against, and does not extend the run further to try to resolve that ambiguity.

\subsection{Where the Weights Changed: A $\Delta W$ Localization}
\label{sec:results-deltaw}

Section~\ref{sec:results-add-robustness} locates reinforcement learning's effect at the level of the weights rather than live inference-time computation; Section~\ref{sec:results-jlens} shows the thought representations themselves are already fixed by the end of the latent curriculum and do not change further under RL. Neither result says what actually changed. Rung-3 and Stage-2 share every LoRA-adapted module and differ only in the 150 GRPO steps run between them, so the resulting delta at each (layer, projection) slot, $\Delta W = \text{scaling} \cdot (B_{\text{new}} A_{\text{new}} - B_{\text{old}} A_{\text{old}})$ in the notation of \citep{hu2021lora}, is a fully isolated, directly inspectable object rather than something inferred indirectly through behavior. We reconstruct $\Delta W$ at all 280 (layer, projection) slots across both checkpoints and, following the weight-level localization approach used elsewhere in interpretability work \citep{meng2022locating}, ask two questions: where does this change concentrate, and does an independently trained RL delta look structurally similar to it.

\textbf{Localization.} We measure concentration as each projection type's share of the total \emph{energy} in $\Delta W$, the sum of squared Frobenius norms rather than raw norms -- the natural unit once the rank analysis below is introduced, since squared Frobenius norm is exactly the sum of squared singular values. By this measure the change is concentrated rather than diffuse: 86.2\% of the total energy sits in the three MLP projections against 13.8\% in the four attention projections, and one projection alone, \texttt{gate\_proj}, accounts for 56.5\% of the whole adapter's change energy -- more than any other single projection type (Table~\ref{tab:deltaw-module}). The layer-wise breakdown (Appendix~\ref{app:deltaw}) shows this concentration is not driven by a handful of outlier layers: every one of the 40 layers contributes a broadly comparable share, with a mild upward drift in the deepest third of the network.

\textbf{Rank structure.} Localization alone does not say whether the change within an implicated projection is a single coherent adjustment or one spread across the LoRA update's full available rank. We answer this with an exact singular value decomposition of $\Delta W$ at all 280 slots (methodology and full detail in Appendix~\ref{app:svd}). Effective rank, the reciprocal participation ratio over each slot's normalized squared singular values, averages 2.5 across all 280 slots (median 2.2, out of a possible 32). The ten largest movers by Frobenius norm are themselves the most concentrated: layer 19's \texttt{down\_proj}, the single largest weight change in the model, carries 89\% of its own energy in one singular direction alone (full ranking in Appendix~\ref{app:svd}). Reinforcement learning's effect on the weights is therefore not only localized to a small set of projections, but low-rank within them: a small number of targeted, coherent adjustments rather than a broad reweighting of the available capacity.

\begin{table}[h]
\centering
\begin{tabular}{lccc}
\toprule
Projection & Branch & \% of total energy in $\Delta W_{\text{Rung-3}}$ & Mean cosine sim.\ vs.\ $\Delta W_{\text{Rung-1}}$ \\
\midrule
\texttt{gate\_proj} & MLP  & 56.5 & 0.040 \\
\texttt{up\_proj}   & MLP  & 18.7 & 0.017 \\
\texttt{down\_proj} & MLP  & 10.9 & 0.022 \\
\texttt{q\_proj}    & Attn & 7.6  & 0.047 \\
\texttt{o\_proj}    & Attn & 4.1  & 0.012 \\
\texttt{k\_proj}    & Attn & 1.3  & 0.042 \\
\texttt{v\_proj}    & Attn & 0.9  & 0.037 \\
\bottomrule
\end{tabular}
\caption{RL-induced weight change by projection type, summed across all 40 layers. \% of total is each projection's share of the summed \emph{squared} Frobenius norm (energy) of $\Delta W_{\text{Rung-3}} = W_{\text{Rung-3}} - W_{\text{Stage-2}}$ across all seven projection types, the same unit used by the effective-rank analysis in \S\ref{sec:results-deltaw} and Appendix~\ref{app:svd}. Mean cosine similarity is against the independently trained $\Delta W_{\text{Rung-1}} = W_{\text{Rung-1-v2}} - W_{\text{SFT}}$ delta, averaged per-slot within each projection type across all 40 layers.}
\label{tab:deltaw-module}
\end{table}

\textbf{Cross-recipe structural comparison.} Rung 1's own adapter is trained under the identical GRPO reward, but starting from the plain SFT checkpoint and reasoning in explicit words, rather than from Stage-2's latent-curriculum checkpoint -- a second, independently produced RL weight delta, reached by a different route and over a different reasoning format. Whole-adapter cosine similarity between $\Delta W_{\text{Rung-3}}$ and $\Delta W_{\text{Rung-1}}$ is $+0.034$ (mean per-slot: $0.031$; median: $0.032$; range across all 280 slots: $[-0.205, 0.240]$). In isolation this is a small number, but not a meaningless one: at rank 16 in a $5120$-dimensional space, two genuinely unrelated deltas would concentrate tightly around zero, so a small value that stays consistently positive across the large majority of slots is a real, if modest, signal rather than noise. That signal is not uniform across depth. Averaged over the first 27 layers, mean per-slot cosine similarity is $+0.052$; over the final 13 layers it falls to $-0.011$, essentially zero. The two independently trained deltas share a faint but consistent family resemblance through most of the network's depth, and diverge specifically in the final third -- the layers closest to committing to a decoded move.

Read together with Sections~\ref{sec:results-add-robustness} and \ref{sec:results-jlens}, this gives the paper's mechanistic claim a physical location rather than only a behavioral signature: RL's effect is not spread evenly through the adapter, it concentrates in the MLP and specifically in \texttt{gate\_proj}, it is low-rank within the projections it touches rather than diffuse, and it is partly, though not fully, a generic consequence of the reward -- shared across two differently-trained recipes through most of the network, but recipe-specific in the layers nearest the output.

\section{Discussion}

\subsection*{Limitations}

Every number in this paper comes from one training run per checkpoint, not an
average across seeds. No result here should be read as more statistically
certain than a single-seed result can be. One thing partially offsets this:
the six causal-battery conditions in Table \ref{tab:causal} are not six copies of the same
measurement, they are six structurally different interventions, and the
pattern they produce -- near-costless under content swaps or randomization,
mildly costly under removal, catastrophic only at exact zero -- repeats
independently on Stage-2 and on Rung-3. A pattern that survives six different
corruptions on two checkpoints from the same training lineage is weaker
evidence than the same number surviving independent seeds, but it is not the
same as having no seed evidence at all. The exact percentages in every table should be treated as point estimates from one training run, not estimates of seed-to-seed variance. The retention-gap confidence intervals in Appendix \ref{app:significance} quantify a different kind of uncertainty — sampling variance over the fixed 100-position harness — and do not substitute for it.

The scope is narrow by design, not by oversight. Every checkpoint is a LoRA
adapter on one 14B-parameter base model, trained in one domain. We make no
claim here about full fine-tuning, frontier-scale models, or reasoning
outside chess.

More consequential than either of those: this paper does not claim the model
plays better chess. Accuracy, exact agreement with Stockfish's own top move,
sits at 9--10\% for every checkpoint in Table \ref{tab:headline}, Rung-3 included.
Reinforcement learning moves legality and removes confabulation; it leaves
accuracy untouched. That is not a gap we hoped would close and quietly
didn't: it lands on the same ceiling an independent chess-RLVR study reports
on a different model and recipe \cite{hwang2025chess}, and our own
explicit-reasoning checkpoint (Rung 1, 52\% legal, the same flat accuracy)
reproduces that ceiling directly before the latent results even begin. Two
independent recipes hitting the same wall is evidence the wall is real, not
evidence of an under-tuned run. Furthermore, a paired centipawn-loss comparison on the 42
boards where both Stage-2 and Rung-3 produced a legal baseline move shows no
meaningful difference (median diff $0$cp, mean diff $19$cp favoring Stage-2),
confirming that the flat exact-match accuracy above is not concealing a
quality gain invisible to a stricter, continuous metric.

The Gumbel control (\ref{sec:method-gumbel}, \ref{sec:results-gumbel}) was included to test whether a working gradient into thought content would produce a stronger policy — the most plausible mechanism by which our deterministic recipe could be missing a better solution. It did not: legality plateaued below the deterministic recipe (50\% vs. 61\%), and confabulation persisted at 15/100 rather than reaching zero. Whether this reflects a genuine absence of usable content in the thoughts or a structurally harder optimization path is unresolved; the straight-through Gumbel estimator is a known-biased gradient approximation. Because the main behavioral and mechanistic results (\ref{sec:results-headline}, \ref{sec:results-add-robustness}, \ref{sec:results-jlens}) do not depend on this run, the ambiguity does not affect our primary claim.

The base-model and SFT-checkpoint comparisons in \ref{sec:results-jlens} (Appendix~\ref{app:basemodel})
carry a caveat of their own: neither checkpoint was ever trained to feed its own hidden
state back in as an input embedding, so their cosine-similarity figures may partly
reflect an unfamiliar-input effect rather than a purely architectural tendency toward
board-invariant representations. The comparison the paper's claim actually rests on,
Stage-2 against Rung-3, is not subject to this caveat, since both checkpoints were
trained on the mechanism directly.

\subsection*{Broader Implications}

The claim above is scoped to one training recipe, not a verdict on latent
reasoning generally: the causal battery does not show that continuous
thoughts are inert, only that in this pipeline the payoff from
reinforcement learning does not require them to be read, in any
content-sensitive way, at inference time. Even scoped this narrowly, it
bears directly on a claim the field treats as closer to settled than it is.
Coconut's own paper frames a continuous thought as able to hold several
candidate next steps at once, an internal scratchpad the model consults
mid-inference \cite{le2024coconut}. A thought vector that costs nothing to
average away, replace with a fixed stand-in, or swap for a different
board's thought entirely is not functioning as a scratchpad in any sense
that would matter to the move the model actually plays.

We are not the first to doubt this, and the field has not settled it
either way. On the skeptical side, a causal and adversarial analysis of
Coconut itself finds its latent tokens behave as "uninterpretable
placeholders" that models lean on for shortcuts rather than faithful
intermediate computation, on ordinary QA and reasoning benchmarks
\cite{zhang2025dolatent}. A separate study reaches the same conclusion from
a different angle, injecting graded noise directly into several
latent-reasoning methods' thought vectors on GSM8K and finding accuracy
survives disruption that should matter if the content were load-bearing
\cite{cui2026supervision} -- the same signature our own noise and
substitute conditions produce. On the other side, a recent interpretability
study reports the opposite in scaled-up models: causal and geometric probes
find intermediate thought vectors carrying genuine, compressed information
about the reasoning steps they stand in for, with the earliest positions
acting as the causally decisive links in the chain \cite{chang2026unlocking}.
A NeurIPS 2025 case study on CODI reports comparable positive structure, an
alternating storage-then-computation cycle recoverable through activation
patching \cite{goyal2025scratchpad}. That specific result has itself since
been challenged on methodological grounds: a follow-up causal-geometric
analysis reruns the same tests against matched controls that lack CODI's
recurrence or curriculum entirely, finds the same alternating pattern in
the controls too, and argues that observable structure alone, whether
decoded, attended to, or geometrically tidy, cannot establish that a
mechanism is doing real work without a causal test run against a matched
baseline \cite{aswal2026observable} -- precisely the design choice, matched
conditions applied identically before and after training rather than read
off one trained checkpoint, our own causal battery was already built
around.

None of this prior work was run in chess, and none of it compares the
identical battery on the same model before and after an RL stage. That gap
is not only a novelty claim: if the field builds monitoring or steering
tools on the premise that a model's latent thoughts hold readable,
load-bearing content, whether that premise holds looks like it depends on
the training recipe and domain, not something safe to assume transfers
wholesale. Our result adds one more point to that map, in a domain and a
training regime this debate has not yet reached, and it lands, again, on
the skeptical side.

\subsection*{Direct Engagement with Switch}
\label{sec:discussion-switch}

The closest prior work to the causal argument in this paper is Switch
\cite{yang2026switch}, a hidden-state-recurrence latent reasoner trained
with GRPO on MATH-500 and GSM8K, on Qwen3-8B. Its causal battery
intervenes on the same class of object we do, the injected hidden state
at each latent position, with three conditions: zeroing it, replacing
it with a random vector of matched norm, and skipping the latent step
entirely. On problems where their baseline model both used the latent
path and answered correctly, zeroing collapses accuracy from 100\% to
33.3\%; the matched-norm random vector costs 9.5 points; skipping the
latent step costs 19.0 points. The ordering, catastrophic only at exact
zero, moderately costly to remove, cheapest under a content swap, is
the same ordering our own battery produces (Table \ref{tab:causal}). Switch treats
this pattern as evidence that the latent step is functionally
load-bearing rather than something the surrounding text could produce
on its own \cite{yang2026switch}. We read a structurally similar
pattern, on our own model, as evidence the model does not depend on
thought content at all. Same shape of result, opposite conclusion.

Two things keep this from being just a difference of opinion. First,
the actual numbers are not identical, only similarly ordered. Our
content-preserving conditions, a fixed averaged vector and independent
random noise alike, cost Rung-3 nothing measurable (58\% baseline
versus 58\% and 60\%); Switch's matched-norm random vector costs a real
9.5 points, on their own post-RL checkpoint. We do not know whether
that gap reflects the math-versus-chess domain, the different base
models, or Switch's narrower diagnostic-subset denominator (problems
the baseline already solved using the latent path, a smaller and more
curated set than our full 100-position harness), and we are not going
to guess. It is a genuine, unresolved point of quantitative
disagreement, not something our design settles.

Second, and more decisively, the two papers disagree about content, not
just about robustness, and that disagreement rests on tools of very
different strength. Switch's evidence for what the latent step contains
is a single logit-lens read at the first latent position, showing the
top candidates growing less concentrated and starting to track features
of the specific problem; their own Limitations section states plainly
that this reading "should not be read as a faithful reconstruction of
the latent reasoning trajectory" \cite{yang2026switch}. Our J-lens
finding (\ref{sec:results-jlens}) is the mirror image, but calibrated: the same instrument
recovers real content at roughly 94\% on ordinary text at the layers we
read, roughly 90 points above the plain logit lens at the same depth,
and only then, pointed at the real thoughts, returns a null (cross-board
cosine $\approx$0.99, no problem-specific content). A quantified null
from an instrument with a known true-positive rate is a stronger claim
than a qualitative positive from an instrument its own authors decline
to vouch for. We do not think Switch is wrong that their latent step
does something; we think the specific claim that it carries
problem-relevant content is not yet established with the same rigor
their own causal-robustness result has. We additionally confirm this null is not an artifact of adapter training: the same
measurement on the raw base model and the SFT checkpoint, neither of which was ever
trained to use this feedback mechanism, returns a lower but still elevated similarity
(0.874 and 0.812 respectively), rising to a stable 0.992 only once the latent
curriculum is applied and remaining unchanged after reinforcement learning
(Appendix~\ref{app:basemodel}) -- a control neither Switch's single logit-lens read nor
our own original diagnostic provided.

One further difference is a matter of design, not interpretation.
Switch's causal battery, like the content-characterization it draws on,
is run on a single post-RL checkpoint; nothing in their tables isolates
what reinforcement learning itself changed about robustness to
disruption. That before/after comparison, the one this paper's central
claim rests on, remains, as far as we can tell, unclaimed elsewhere,
chess included.

\subsection*{Future work}

Section~\ref{sec:results-deltaw} shows both where the RL-induced weight
change concentrates (predominantly the MLP projections, and
\texttt{gate\_proj} specifically) and that it is low-rank within those
locations -- a small number of coherent directions rather than a diffuse
reweighting of the LoRA update's full capacity. Localizing and
characterizing the change this precisely is not the same as establishing
what it computes. The natural next step is causal, not descriptive: patch
the dominant singular directions identified in Appendix~\ref{app:svd},
most directly layer 19's \texttt{down\_proj} direction, responsible for
89\% of that slot's own change energy, into the pre-RL Stage-2 checkpoint
and observe whether legal-move rate moves toward Rung-3's, or ablate them
from Rung-3 and observe whether it moves back toward Stage-2's. A result
either way would connect this section's weight-space description to the
causal battery's behavioral one directly, rather than leaving the two
linked only by both concerning the same reinforcement-learning stage.

\section{Related work}

\subsection*{Latent reasoning}

Latent (or silent) reasoning originates with Coconut \cite{le2024coconut}
(\ref{sec:background-latent}), which established both the mechanism used throughout this paper
and the specific claim we test: that a continuous thought can encode more
than one candidate reasoning step at once, functioning as an active
scratchpad the model consults during inference. Whether that claim
survives reinforcement learning has, until now, been tested only in math
and logic settings, and the reported outcome has consistently been
negative rather than mixed. Ouro, the current looped-language-model state
of the art, reports that RLVR training on top of its supervised checkpoint
did not yield significant gains \cite{zhu2025ouro}; a separate study
applying GRPO directly to a Coconut-style model reports the same null
result, tracing it to GRPO's update mechanism having no intermediate
tokens to act on \cite{ozeren2025rl}. Both papers treat this as a
limitation of the recipe, not of the domain; this paper's contribution is
showing the recipe does work, in a domain neither tested.

\subsection*{Reinforcement learning for chess}

RLVR applied to chess, independent of any latent-reasoning question,
plateaus at a real but modest ceiling. Hwang et al. train an LLM to
reason explicitly about chess positions under RLVR and report that move
quality improves only up to a limit set by the pretrained model's own
chess understanding, a gap reinforcement learning alone could not close
\cite{hwang2025chess}; their paper is also the one the field points to
for establishing that a working RLVR result in chess remained an open
question at the time. Rung 1 in this paper, GRPO applied to explicit
chain-of-thought reasoning with a gated legality reward, reaches a
comparable ceiling (52\% legal, flat accuracy) via an independently built
training pipeline and reward design, functioning as an unplanned but
direct replication of their finding. The behavioral gain this paper
reports, legality and confabulation, sits on an axis their
accuracy-focused evaluation does not measure, which is why Rung 1
replicating their ceiling and Rung 3 exceeding it on a different axis are
not in tension with each other.

\subsection*{Mechanistic interpretability and causal intervention}

The causal battery in \ref{sec:method-causal-suite} and the J-lens analysis in \ref{sec:results-jlens} both draw on
an established methodological lineage rather than inventing intervention
or reading techniques from scratch. Causal mediation analysis
\cite{vig2020investigating} and causal tracing \cite{meng2022locating}
(\ref{sec:background-causal}) establish the general principle that a representation's role must
be tested by intervention, not inferred from correlation alone; a
systematic comparison across activation-patching methodologies
\cite{zhang2024best} further shows that the specific choice of corruption
and replacement value can itself change the interpretability conclusion
drawn from a fixed model and dataset, the direct motivation for running
six intervention types rather than one. We follow that lesson at the
level of the intervention (\ref{sec:method-causal-suite}) and again at the level of the reading
tool, calibrating J-lens against known content before trusting its null
result on the latent thoughts (\ref{sec:results-jlens}), rather than treating either
instrument's output as self-certifying.

\subsection*{Switch: closest prior art}

The closest work to this paper's causal argument is Switch
\cite{yang2026switch}, a hidden-state-recurrence latent reasoner in the
same Coconut family as the mechanism used throughout this paper (\ref{sec:background-latent}),
extended with a pair of learned boundary tokens, \texttt{<swi>} and
\texttt{</swi>}, that mark where a latent block begins and ends. Because
those boundaries are ordinary discrete tokens, GRPO's policy ratio is
well-defined at every text position around the latent block, one way of
resolving the fact that latent positions have no policy density; our own
reward design resolves the same problem differently, by never scoring the
thoughts directly and rewarding only the decoded move (\ref{sec:method-training}). On a
Qwen3-8B base, Switch reaches 79.3\% on MATH-500 and 89.2\% on GSM8K,
above every Coconut-style baseline tested at the same scale. Chess is not
among the domains they evaluate.

Switch does report genuine before/after-reinforcement-learning
comparisons for two quantities: the model's calibration for when to
invoke the latent path at all ($p(\texttt{<swi>})$ falls from 0.85 to
0.48 after RL, meaning the trained policy invokes latent reasoning more
selectively rather than reflexively) and how linearly decodable that
switch decision is from late-layer activations (91.9\% to 88.4\%, a small
drop). Both comparisons speak to the switching decision, not to what the
latent step itself carries or how sensitive the model is to disrupting
it. Their own causal battery, the zero/random-norm/skip conditions
engaged with directly in \ref{sec:discussion-switch}, shows no equivalent before/after split in
their reported tables and, as far as their text indicates, is run on the
post-RL checkpoint alone.

A dedicated search for prior work running the identical causal battery on
the identical model both before and after an RL stage, the design this
paper's central claim rests on, returned nothing, in chess or elsewhere,
including Switch. Chess itself remains untested by every paper in this
literature we are aware of. Both gaps are what this paper fills; the full
technical comparison to Switch's own causal-battery numbers and its
conflicting interpretation is in \ref{sec:discussion-switch}.

\section{Conclusion}

This paper asked whether the latent, silent thoughts a language model
produces during training function as an actively consulted scratchpad, or
as scaffolding that shapes the model's weights during training without
needing to be read at inference time. On a chess-playing model trained
through a staged latent-thought curriculum followed by reinforcement
learning, we find real behavioral gains: legal-move rate climbs
monotonically to 61\% from a 48\% pre-RL baseline, and checkmate
confabulation, confidently declaring checkmate on positions where none
exists, is eliminated entirely, independently, across two different
reasoning formats (the Confabulation Cutoff). That elimination is not
incidental: the false-checkmate explanations that do occur earlier in
training are themselves stereotyped, reusing near-identical wording
across unrelated boards (Templated Confabulation), evidence of a fixed,
memorized script rather than board-specific reasoning gone wrong. A six-condition causal battery, run on the same model before and after reinforcement learning, locates where the behavioral gain comes from: neither checkpoint relies on the specific content of its latent thoughts, but the post-RL checkpoint retains a significantly larger share of its own baseline competence under total signal loss, a result that holds even after correcting for testing all five non-baseline conditions as a family. A replication of the post-RL checkpoint's own robustness pattern at ten times the sample size confirms this trend extends further than the original sample could establish on its own: removing the thoughts, with or without restoring the sequence's length, costs the post-RL checkpoint a small but statistically real amount, while substituting or randomizing their content costs nothing measurable at any sample size tested. Reinforcement
learning, in this setting, does not teach the model to consult a richer
internal scratchpad; it reshapes the weights themselves, leaving behavior
robust regardless of what occupies the silent thoughts. Accuracy against
Stockfish's own best move stays flat throughout, so this is a legality
and confabulation result, not a claim that the model plays stronger
chess. Taken together, these findings push back against latent
reasoning's founding scratchpad assumption, in a domain, chess, where
reinforcement learning over latent reasoning has otherwise been reported
to fail.

\appendix
\section{Appendix}
\label{app:main}

\subsection{Reward Function}
\label{app:reward}

The reward used for Rung 1 and Rung 3 alike is a single gated function,
unchanged between the two:
$$
R(\text{response}) =
\begin{cases}
-1.0 & \text{output unparseable, malformed, illegal, or a false CHECKMATE claim} \\[4pt]
+1.0 & \text{response correctly identifies CHECKMATE} \\[4pt]
\exp(-\text{cp\_loss}/120) & \text{otherwise (legal, non-terminal move)}
\end{cases}
$$
where $\text{cp\_loss} = \max(0,\ \text{eval}_{\text{best}} - \text{eval}_{\text{move}})$.
Stockfish is never called on a gate failure; the played move is pushed and
the resulting position is scored from the mover's point of view at
\texttt{Limit(depth=12, time=10.0)}, \texttt{Threads=1} (for determinism),
\texttt{Hash=256MB}, \texttt{mate\_score=1500}. A single persistent engine
process is reused across the run; if a legal move's evaluation fails twice
consecutively, the position falls back to a flat reward of 0.1 rather than
crashing the run. The exponential decay means a 50cp loss still scores
$\approx$0.66, a 200cp loss $\approx$0.19, and a 500cp loss $\approx$0.02:
quality is scored on a continuous, forgiving scale once a move clears the
legality gate, rather than as a second pass/fail threshold. An earlier,
rejected design summed a legality term and a quality term instead of
gating one behind the other; every result in this paper uses the gated
version, for the reasons given in \ref{sec:background-rlvr} and \ref{sec:method-training}.

\subsection{Frozen Evaluation Harness}
\label{app:harness}

The 100-position evaluation set is a fixed split (seed 3407) of a
4{,}500-row pool restricted to positions with a precomputed Stockfish
best-move evaluation, itself drawn from the 11{,}647 canonical rows of the
full dataset (\ref{sec:method-training}); the remaining 4{,}400 rows of that pool are used for
GRPO training, never evaluation. The prompt is manual ChatML, byte-identical
across SFT, GRPO, and evaluation:

\begin{quote}
\ttfamily\small
You are a Chess Grandmaster. Given a FEN string, analyze the position and determine the best move.\\[4pt]
Format:\\
1. Reason inside \textless thinking\textgreater\ tags using [KING SAFETY], [CHECKS], [CAPTURES \& TRADES], [THREATS], [IMPROVEMENT] sections.\\
2. Output the move in UCI format inside \textless output\textgreater\ tags.
\end{quote}

\noindent \texttt{apply\_chat\_template} is never used, to avoid drifting
off the exact byte format the model was trained on. Decoding is greedy
(temperature 0) for every checkpoint. Latent checkpoints (Stage 1 through
3, Rung-3, and the Gumbel run) are additionally run at batch size one,
since bf16 outputs can depend weakly on batch composition and batch size
one removes that variable entirely. All legality and checkmate scoring
uses \texttt{python-chess} directly against the position's FEN. The four
metrics, Format, Legal, Accuracy, False Mate, are defined identically to
\ref{sec:method-frozen-eval-harness} for every checkpoint and every causal-battery condition; the harness
itself was frozen after Rung 1 and never modified afterward.

\subsection{Training Curves}
\label{app:curves}

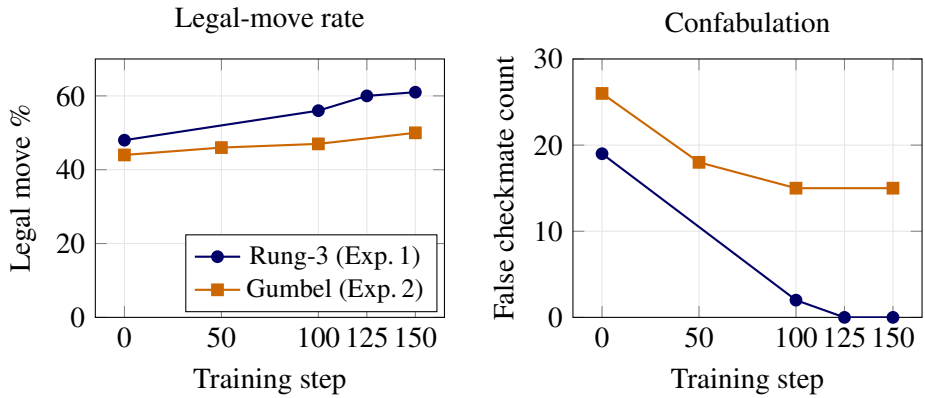
\begin{figure}[H]
\centering
\begin{tikzpicture}
\begin{groupplot}[
    group style={group size=2 by 1, horizontal sep=1.7cm},
    width=6.2cm, height=5cm,
    xlabel={Training step},
    xtick={0,50,100,125,150},
    legend style={font=\small, at={(0.98,0.02)}, anchor=south east},
    grid=major,
    grid style={gray!20},
]
\nextgroupplot[ylabel={Legal move \%}, ymin=0, ymax=70, title={Legal-move rate}]
\addplot[color=blue!40!black, mark=*, thick] coordinates {(0,48) (100,56) (125,60) (150,61)};
\addlegendentry{Rung-3 (Exp.\,1)}
\addplot[color=orange!80!black, mark=square*, thick] coordinates {(0,44) (50,46) (100,47) (150,50)};
\addlegendentry{Gumbel (Exp.\,2)}

\nextgroupplot[ylabel={False checkmate count}, ymin=0, ymax=30, title={Confabulation}]
\addplot[color=blue!40!black, mark=*, thick] coordinates {(0,19) (100,2) (125,0) (150,0)};
\addplot[color=orange!80!black, mark=square*, thick] coordinates {(0,26) (50,18) (100,15) (150,15)};
\end{groupplot}
\end{tikzpicture}
\caption{Checkpoint-level training trajectories, from
\texttt{rung3\_history\_resume.json} and \texttt{rung3\_gumbel\_history.json}.
Left: legal-move rate. Right: false-checkmate count. Rung-3's gap between
step 0 and step 100 is not in the recovered history file and is not
plotted as a continuous line over that interval.}
\label{fig:curves}
\end{figure}

Table~\ref{tab:curves} and Figure~\ref{fig:curves} report the
checkpoint-level trajectory for both 150-step RL runs, taken directly from
the recorded per-checkpoint harness evaluations. Rung 3's step-0 point is
Stage-2's own harness score, its effective initialization value, not a
logged training-step read; the run's steps 0--100 window (from the
pre-crash history file, \ref{sec:method-training}) is not available and is not plotted, only
the post-recovery steps 100, 125, and 150. Finer-grained per-step loss and
KL logs exist for both runs but are not reproduced here. One qualitative
note from those logs is worth stating plainly: Rung 3's KL stayed alive
throughout its run, in the range $0.0006$--$0.006$, in contrast to Rung
1's flat $\sim$1.19 (\ref{sec:results-headline}), consistent with the policy actually moving
under Rung 3 rather than sitting near its initialization.

\begin{table}[H]
\centering
\begin{tabular}{lccccc}
\toprule
Run & Step & Format\% & Legal\% & Acc.\% & False Mate \\
\midrule
Rung 3 & 0 (= Stage-2) & 100 & 48 & 10 & 19 \\
Rung 3 & 100            & 100 & 56 & 9  & 2  \\
Rung 3 & 125            & 100 & 60 & 9  & 0  \\
Rung 3 & 150 (final)    & 100 & 61 & 9  & 0  \\
\midrule
Gumbel & 0               & 100 & 44 & 8 & 26 \\
Gumbel & 50               & 100 & 46 & 8 & 18 \\
Gumbel & 100              & 100 & 47 & 8 & 15 \\
Gumbel & 150 (final)      & 100 & 50 & 9 & 15 \\
\bottomrule
\end{tabular}
\caption{Checkpoint-level trajectory for both 150-step RL runs. Format
compliance holds at 100\% throughout both. Rung 3's confabulation count
falls from 19 to 2 by step 100 and reaches 0 by step 125, not in a single
jump; the Gumbel run's count plateaus at 15 from step 100 onward and does
not reach zero within the matched budget (\S4.6).}
\label{tab:curves}
\end{table}

\subsection{Board-Level Qualitative Example}
\label{app:board}

Figure~\ref{fig:templated_confabulation} shows one board's baseline completion
in full. The same board, FEN
\texttt{\seqsplit{r3r1k1/pp2R1Bp/3bB1p1/2p1n3/4p1P1/7P/PPP1K3/7R} b - - 0 23}, is
useful as a worked example of what a genuinely hard position looks like
across the full causal battery, since it is not actually checkmate: Black
has a legal escape, a quiet knight move from e5 to f7 that blocks the
checking bishop on e6. Every non-zero condition run on this position
fails to find it, in one of two ways. Four conditions (baseline and three
others) hallucinate checkmate, each producing a completion in the same
templated register documented in \ref{sec:results-tempconfab}, varying only in which piece and
square is named as delivering the mate. The fifth, noise, does not
hallucinate but instead plays an illegal move, g8h8, which walks the king
directly into the g7 bishop's line of check. No condition run on this
board finds the real defense. We do not read this as evidence against any
claim in the paper; it is offered as a concrete illustration of what
Templated Confabulation and the causal battery's failure modes actually
look like on a single position, rather than only in aggregate.

\subsection{Pairwise Exact-Move-Match Churn Analysis}
\label{app:churn}

Aggregate legal-move rate can hide churn at the level of individual
positions: two conditions can post the same pass rate while disagreeing
about which specific moves are legal. To check this, every pair of
conditions is compared position-by-position on the frozen 100-position
harness, counting an exact match only when both conditions produce the
identical legal move; an unparseable completion on either side always
counts as a mismatch, never as an excluded row. An earlier version of
this analysis called \texttt{.dropna()} before comparing, which silently
inflated the match rate for any condition with many unparseable
completions; the zero condition's true pairwise match rate against
baseline is 14\%, not the 87.5\% the buggy summary line originally
reported. All numbers below use the corrected, non-dropping comparison.

\begin{table}[H]
\centering
\begin{tabular}{lcc}
\toprule
Condition pair & Exact-move match & Flips (of 100) \\
\midrule
Baseline vs.\ Substitute & 92\% & 2 \\
Baseline vs.\ Noise      & 91\% & 2 \\
Substitute vs.\ Noise    & 89\% & 4 \\
Baseline vs.\ Ablate     & 82\% & 9 \\
Substitute vs.\ Ablate   & 80\% & 9 \\
Ablate vs.\ Noise        & 83\% & 9 \\
Baseline vs.\ Zero       & 14\% & 86 \\
\bottomrule
\end{tabular}
\caption{Pairwise exact-move-match agreement, Rung-3 (post-RL), all
condition pairs run on the frozen 100-position harness. Baseline,
substitute, and noise form a tight cluster (89--92\%); ablate sits a
consistent step apart from all three (80--83\%), a small but repeatable
gap; zero stands alone. The corresponding Stage-2 (pre-RL) baseline-vs-noise
figure is 82\%, against Rung-3's 91\% here (\ref{sec:results-add-robustness}); a full Stage-2
pairwise breakdown analogous to this table was not reconstructed for this
draft.}
\label{tab:pairwise-rung3}
\end{table}

A follow-up test on the ablate gap, padding the ablated sequence back to
baseline length with the model's own pad-token embedding
(lenmatch-ablate), did not close it: lenmatch-ablate scores lower than
plain ablate, not higher, because the pad token's entire trained role is
to be masked out of attention, so forcing the model to attend to it
introduces its own out-of-distribution effect rather than isolating
sequence length cleanly. The ablate gap is reported in the main text as directionally consistent with the paper's central claim but not individually significant at n=100 (Appendix \ref{app:significance}), rather than pursued with a further condition.

\subsection{Templated Confabulation: Cross-Board Text Similarity Methodology}
\label{app:template}

Cross-board similarity for every false-checkmate completion in the
Stage-2 causal battery is computed with Python's \texttt{difflib}
(\texttt{SequenceMatcher.ratio()}) over the completion text, comparing
every pair of false-checkmate completions that come from \emph{different}
boards. An earlier pass paired same-board completions from different
causal conditions against each other by mistake, which artificially
inflated the similarity figures for both groups; excluding those pairs
and keeping only genuinely cross-board comparisons barely moved either
number, evidence the pattern is real rather than a counting artifact. The
$n=96$ pooled completions are not evenly drawn from the six causal
conditions:

\begin{table}[H]
\centering
\begin{tabular}{lc}
\toprule
Condition & False-checkmate completions contributed \\
\midrule
Baseline & 19 \\
Substitute & 20 \\
Noise & 17 \\
Ablate & 22 \\
Lenmatch-Ablate & 18 \\
Zero & 0 \\
\midrule
Total & 96 \\
\bottomrule
\end{tabular}
\caption{Composition of the $n=96$ false-checkmate completions pooled
across Table~\ref{tab:causal}'s six Stage-2 conditions for the
cross-board similarity analysis (\ref{sec:results-tempconfab}). Zero contributes none, since its
own false-mate count is 0 (Table~\ref{tab:causal}).}
\label{tab:template-composition}
\end{table}

The comparison group of real, distinct move explanations (0.27 average
similarity, against 0.68 for the false-checkmate group) is sampled across
different boards under the same five-section template, so the gap between
the two groups reflects content, not the shared formatting both groups
inherit from the dataset's fixed output structure (\ref{sec:method-training}).

\subsection{Significance of the Causal Battery}
\label{app:significance}

Table~\ref{tab:significance} reports the full statistical detail underlying the
retention-gap analysis in \ref{sec:results-add-robustness}, across all five non-baseline
conditions: within-checkpoint McNemar tests on the paired baseline-vs-condition
outcomes, and a bootstrap retention-ratio gap between checkpoints (10{,}000
resamples, board-level pairing preserved across checkpoints in every resample),
defined as each checkpoint's own condition legal rate divided by its own baseline
legal rate, Rung-3's retained fraction minus Stage-2's. Because five conditions are
tested from the same battery, we report both the uncorrected 95\% CI and a
Bonferroni-corrected CI (per-test $\alpha=0.01$) for the family.

\begin{table}[h]
\centering
\footnotesize
\begin{tabular}{lccccc}
\toprule
Condition & Stage-2 McNemar $p$ & Rung-3 McNemar $p$ & Retention gap & 95\% CI & Bonferroni CI ($n{=}5$) \\
\midrule
Substitute       & 1.00   & 1.00   & $+0.000$ & [$-0.102$, $0.091$] & [$-0.136$, $0.121$] \\
Noise            & 1.00   & 0.50   & $+0.034$ & [$-0.102$, $0.162$] & [$-0.150$, $0.204$] \\
Ablate           & 0.39   & 1.00   & $+0.066$ & [$-0.084$, $0.209$] & [$-0.132$, $0.256$] \\
Lenmatch-Ablate  & 0.023  & 0.27   & $+0.101$ & [$-0.052$, $0.256$] & [$-0.105$, $0.309$] \\
Zero             & $<0.001$ & $<0.001$ & $+0.134$ & [$0.050$, $0.231$] & [$0.031$, $0.263$] \\
\bottomrule
\end{tabular}
\caption{Full significance testing for all five non-baseline conditions in
Table~\ref{tab:causal}. McNemar $p$-values test whether the baseline-vs-condition
outcome flip is asymmetric within a single checkpoint. Retention gap is each
checkpoint's own condition legal rate divided by its own baseline legal rate,
Rung-3's retained fraction minus Stage-2's; the 95\% CI is the uncorrected
bootstrap interval, and the Bonferroni CI applies a family-wise correction for
testing all five conditions together (per-test $\alpha=0.01$). Only Zero clears
significance, under both the uncorrected and the corrected threshold; the
remaining four conditions trend in the same direction with increasing severity but
do not individually reach significance at $n=100$.}
\label{tab:significance}
\end{table}

One nuance is worth disclosing directly. The retention-ratio framing above
normalizes each checkpoint's condition-legal count by its own baseline count
before comparing checkpoints, which is the correct comparison when the two
checkpoints start from different baseline legality (48\% pre-RL vs.\ 58\% within
this appendix's internally consistent baseline, \ref{sec:results-add-robustness}).
A different, unnormalized framing -- counting the raw number of boards that flip
from legal to illegal under each condition -- tells a less clean story: under
Zero, Stage-2 loses 47 of its 100 boards to illegality and Rung-3 loses 49, a
difference in the opposite direction, and a paired Wilcoxon test on this raw-drop
framing does not reach significance for any of the three most severe conditions
(Ablate $p=0.22$, Lenmatch-Ablate $p=0.17$, Zero $p=0.66$, one-sided). We report
both framings rather than only the one that supports the paper's claim: Rung-3
starts Zero with more legal boards to lose (58 versus 48), so it can lose more of
them in absolute count while still retaining a significantly larger proportion of
its own baseline competence. Which framing is the right one depends on the
question being asked -- proportional robustness (the paper's claim) versus
absolute board-level stability -- and we consider the retention-ratio framing the
more defensible measure of the two for comparing checkpoints with different
starting legality, but the raw-count result is included here so a reader can judge
the tension directly rather than encountering only the framing that favors the
paper's argument.

\subsection{Large-Sample Replication of the Rung-3 Causal Battery ($n=1000$)}
\label{app:n1000}

The retention-gap analysis above tests all six causal conditions at $n=100$, the sample size used throughout this paper's frozen harness (\S\ref{sec:method-frozen-eval-harness}). Only Zero reaches significance there; the remaining four conditions trend in the predicted severity order without individually clearing the bar. A null result at $n=100$ does not distinguish a genuinely absent effect from one too small for 100 positions to resolve. To check which explanation holds, we replicate Rung-3's own side of the battery, all six conditions, at $n=1000$, an order of magnitude more positions. Stage-2 is not rerun at this scale, so this replication cannot update the paired Stage-2-versus-Rung-3 comparison in Table~\ref{tab:causal}; it tests only whether Rung-3's own point estimates and single-checkpoint significance calls hold up under more sampling power. The evaluation set is a strict superset of the original: the same seed-3407 shuffle used throughout this paper means the first 100 of these 1000 positions are the identical positions used everywhere else in this paper, not a fresh, unrelated sample.

\begin{table}[H]
\centering
\begin{tabular}{lcccc}
\toprule
Condition & Format \% & Legal \% & Accuracy \% & False Mate \\
\midrule
Baseline          & 100.0 & 56.7 & 10.2 & 5  \\
Substitute        & 100.0 & 56.5 & 10.2 & 5  \\
Noise             & 100.0 & 55.5 & 9.9  & 2  \\
Ablate            & 100.0 & 54.7 & 9.7  & 10 \\
Lenmatch-Ablate   & 83.3  & 46.1 & 8.9  & 4  \\
Zero              & 19.0  & 11.3 & 1.6  & 1  \\
\bottomrule
\end{tabular}
\caption{Rung-3, all six causal conditions, $n=1000$. Point estimates are close to their $n=100$ counterparts in Table~\ref{tab:causal} for every condition except Lenmatch-Ablate, discussed below.}
\label{tab:n1000-summary}
\end{table}

Two results change with the larger sample. First, every non-zero condition read 0 false-checkmate claims at $n=100$ (Table~\ref{tab:causal}'s Rung-3 column). At $n=1000$, Baseline and Substitute both show 5 (0.5\%), Ablate shows 10 (1.0\%), Noise shows 2 (0.2\%), and Lenmatch-Ablate shows 4 (0.4\%). A rate this low is consistent with a true confabulation rate near zero being unobservable at $n=100$ from insufficient sampling alone, rather than evidence against the Confabulation Cutoff (\S\ref{sec:results-tempconfab}): a 28\% (SFT) to roughly 0.5--1\% (Rung-3) reduction is still better than a 97\% collapse, and the frozen 100-position headline harness (Table~\ref{tab:headline}), evaluated independently of this causal-battery script, is unaffected and continues to read exactly 0.

Second, we test each condition's legal-rate gap against Baseline with the same paired McNemar methodology used for the $n=100$ within-checkpoint tests in Table~\ref{tab:significance}, now applied to the per-position discordant-pair counts at $n=1000$ (Table~\ref{tab:n1000-significance}). Substitute and Noise remain statistically indistinguishable from Baseline, consistent with our central claim that content does not matter to either checkpoint. Ablate, directionally consistent but short of significance at $n=100$ (Table~\ref{tab:significance}), now clears it ($p=0.021$): the larger sample resolves this specific ambiguity in favor of a real, if modest, effect, rather than against one. Zero and Lenmatch-Ablate both reach significance overwhelmingly. Read together, the five conditions now cross into significance in exactly the same severity order their point estimates were already trending in -- Substitute and Noise remain flat and non-significant; Ablate, Lenmatch-Ablate, and Zero, in increasing order of severity, are all significant -- the same monotonic pattern reported for the point estimates alone in \ref{sec:results-add-robustness}, now visible in the significance calls themselves and not only their direction.

\begin{table}[H]
\centering
\begin{tabular}{lcccc}
\toprule
Condition vs.\ Baseline & Legal-\% gap & Discordant pairs ($b$, $c$) & McNemar $p$ \\
\midrule
Substitute       & $-0.2$  & 11, 9   & 0.824 \\
Noise            & $-1.2$  & 33, 21  & 0.134 \\
Ablate           & $-2.0$  & 44, 24  & 0.021 \\
Lenmatch-Ablate  & $-10.6$ & 110, 4  & $6.7\times10^{-28}$ \\
Zero             & $-45.4$ & 456, 2  & $2.8\times10^{-133}$ \\
\bottomrule
\end{tabular}
\caption{Paired McNemar tests, Rung-3 Baseline against each condition, $n=1000$ positions with complete data across all six conditions, methodologically identical to the $n=100$ within-checkpoint tests in Table~\ref{tab:significance}. $b$ is the count of positions legal under Baseline and illegal under the condition; $c$ is the reverse. Substitute and Noise remain indistinguishable from Baseline; Ablate now reaches significance; Zero and Lenmatch-Ablate both reach it overwhelmingly.}
\label{tab:n1000-significance}
\end{table}

Lenmatch-Ablate's move from a directionally-consistent but non-significant $n=100$ result (Table~\ref{tab:significance}) to a decisively significant one at $n=1000$ motivates a comparison the original battery did not make directly: Lenmatch-Ablate against plain Ablate, rather than either against Baseline. At $n=1000$, Ablate (54.7\% legal) significantly exceeds Lenmatch-Ablate (46.1\% legal), tested with the same paired McNemar methodology used throughout this appendix (113 of 1000 positions legal under Ablate and illegal under Lenmatch-Ablate, against 27 the reverse; $p=1.04\times10^{-13}$). Filling the four thought positions with the model's own pad-token embedding is not a content-neutral way to restore sequence length, as Appendix~\ref{app:churn} already notes qualitatively; at this sample size the cost of doing so is not only real but larger than the cost of removing the positions outright. Forcing attention onto a token whose entire trained role is to be masked out introduces its own out-of-distribution cost, on top of, not instead of, whatever cost removing the thoughts carries on its own.

\subsection{Board Invariance Across the Training Pipeline}
\label{app:basemodel}

Table~\ref{tab:basemodel} reports cross-board cosine similarity (\ref{sec:results-jlens}) for all four
checkpoints in the training progression, at both the four-board sample used in the
original diagnostic and the full 100-position harness. All values are computed by the
same procedure: capturing each thought vector during the model's own silent
forward-feedback loop, then averaging pairwise cosine similarity across all board pairs
for a given thought slot, then averaging across the four slots. The base model and SFT
checkpoint were never trained to perform this specific feedback operation, so their
values should be read alongside the caveat in \ref{sec:results-jlens}; Stage-2 and Rung-3 were both
trained on it directly.

\begin{table}[h]
\centering
\begin{tabular}{lccccc}
\toprule
Checkpoint & Thought 0 & Thought 1 & Thought 2 & Thought 3 & Mean ($n{=}100$) \\
\midrule
Base (no adapter) & 0.922 & 0.933 & 0.953 & 0.689 & 0.874 \\
SFT               & 0.919 & 0.855 & 0.868 & 0.608 & 0.812 \\
Stage-2           & 0.989 & 0.991 & 0.993 & 0.996 & 0.992 \\
Rung-3            & 0.989 & 0.991 & 0.993 & 0.996 & 0.992 \\
\bottomrule
\end{tabular}
\caption{Cross-board cosine similarity per thought slot, all four checkpoints,
$n=100$ boards. Stage-2 and Rung-3 are trained on the feedback mechanism being
measured; Base and SFT are not (\ref{sec:results-jlens}). At $n=4$, the sample size used in the original
diagnostic, the same ordering holds: Base 0.938, SFT 0.845, Stage-2 0.994, Rung-3
0.994 -- confirming the $n=100$ estimates are consistent with the originally reported
$\approx 0.99$ figure rather than an artifact of sample size.}
\label{tab:basemodel}
\end{table}

\subsection{Layer-wise $\Delta W$ Norms and Cross-Recipe Similarity}
\label{app:deltaw}

Table~\ref{tab:deltaw-layerwise} reports the full layer-by-layer detail underlying
Section~\ref{sec:results-deltaw}: the summed Frobenius norm of $\Delta W$ at each of the
40 layers (summed across all seven projection types at that layer) for both the
Rung-3 delta ($\Delta W_{\text{Rung-3}} = W_{\text{Rung-3}} - W_{\text{Stage-2}}$) and the
Rung-1 delta ($\Delta W_{\text{Rung-1}} = W_{\text{Rung-1-v2}} - W_{\text{SFT}}$), alongside
the mean per-slot cosine similarity between the two deltas at that layer.

\begin{table}[h]
\centering
\footnotesize
\begin{tabular}{cccc|cccc}
\toprule
L & Norm R3 & Norm R1 & Cos & L & Norm R3 & Norm R1 & Cos \\
\midrule
0  & 0.0854 & 0.1839 & 0.0124 & 20 & 0.0954 & 0.2114 & 0.0750 \\
1  & 0.0972 & 0.2087 & 0.0584 & 21 & 0.0980 & 0.2127 & 0.0611 \\
2  & 0.0803 & 0.1585 & 0.0223 & 22 & 0.0992 & 0.2222 & 0.0777 \\
3  & 0.0852 & 0.1809 & 0.0249 & 23 & 0.0999 & 0.2426 & 0.0931 \\
4  & 0.0834 & 0.2030 & 0.0643 & 24 & 0.0994 & 0.2154 & 0.0847 \\
5  & 0.0843 & 0.1877 & 0.0492 & 25 & 0.0985 & 0.2196 & 0.0878 \\
6  & 0.1002 & 0.2218 & 0.0279 & 26 & 0.1013 & 0.2189 & 0.0959 \\
7  & 0.0910 & 0.1866 & 0.0192 & 27 & 0.1045 & 0.2349 & -0.0083 \\
8  & 0.0769 & 0.1605 & 0.0434 & 28 & 0.1143 & 0.2432 & -0.0115 \\
9  & 0.0733 & 0.1575 & 0.0439 & 29 & 0.0995 & 0.2117 & 0.0046 \\
10 & 0.0741 & 0.1645 & 0.0542 & 30 & 0.1033 & 0.2329 & 0.0398 \\
11 & 0.0757 & 0.1747 & 0.0575 & 31 & 0.1021 & 0.2155 & -0.0501 \\
12 & 0.0834 & 0.1854 & 0.0197 & 32 & 0.1089 & 0.2278 & -0.0238 \\
13 & 0.0840 & 0.1853 & 0.0192 & 33 & 0.1024 & 0.2018 & -0.0250 \\
14 & 0.0815 & 0.1846 & 0.0435 & 34 & 0.1045 & 0.2169 & -0.0187 \\
15 & 0.0859 & 0.2019 & 0.0401 & 35 & 0.1162 & 0.2298 & -0.0427 \\
16 & 0.0858 & 0.1989 & 0.0443 & 36 & 0.1060 & 0.2157 & -0.0008 \\
17 & 0.0871 & 0.2044 & 0.0574 & 37 & 0.1146 & 0.2278 & -0.0109 \\
18 & 0.0927 & 0.2091 & 0.0519 & 38 & 0.1305 & 0.2468 & 0.0042 \\
19 & 0.1200 & 0.2641 & 0.0616 & 39 & 0.1166 & 0.2394 & -0.0031 \\
\bottomrule
\end{tabular}
\caption{Layer-wise $\Delta W$ Frobenius norm (summed across all seven projections at that layer) and mean per-slot cosine similarity between the Rung-3 and Rung-1 deltas, all 40 layers. The first 27 layers average $+0.052$ cosine similarity; the final 13 average $-0.011$, essentially zero -- the depth split reported in Section~\ref{sec:results-deltaw}.}
\label{tab:deltaw-layerwise}
\end{table}

\subsection{Singular Value Decomposition of $\Delta W$}
\label{app:svd}

Section~\ref{sec:results-deltaw} reports that $\Delta W_{\text{Rung-3}}$ is low-rank within the projections it concentrates in. Computing this exactly, rather than approximately, exploits the low-rank structure of the LoRA parameterization itself \citep{hu2021lora}: since $\Delta W$ at a given slot is $\text{scaling}_3 B_3 A_3 - \text{scaling}_2 B_2 A_2$, a difference of two rank-16 matrices, it has rank at most 32 regardless of the underlying projection's dimension. Rather than forming this large dense matrix and decomposing it directly, we concatenate $[\,\text{scaling}_3 B_3 \mid -\text{scaling}_2 B_2\,]$ and $[A_3;\,A_2]$, reduce each to an orthonormal basis via QR decomposition, and take the singular value decomposition of the resulting $32\times32$ core matrix. This recovers $\Delta W$'s true singular values and vectors exactly, at the cost of one small SVD per slot rather than one on a matrix with tens of millions of entries.

Effective rank at each slot is the reciprocal participation ratio $1/\sum_i p_i^2$, where $p_i = \sigma_i^2 / \sum_j \sigma_j^2$ is each singular value's share of the slot's total energy: 1.0 for a purely rank-one change, up to 32 for a change spread evenly across every available direction. Table~\ref{tab:svd-top10} reports the ten largest movers by Frobenius norm together with their effective rank; all ten sit well below the maximum, and the single largest mover in the entire model, layer 19's \texttt{down\_proj}, is also nearly the most rank-one of any slot in this table.

\begin{table}[H]
\centering
\footnotesize
\begin{tabular}{ccccc}
\toprule
Layer & Module & Frobenius norm & Top-1 energy share & Effective rank \\
\midrule
19 & \texttt{down\_proj} & 0.0437 & 89.2\% & 1.25 \\
28 & \texttt{gate\_proj}  & 0.0410 & 74.0\% & 1.63 \\
7  & \texttt{gate\_proj}  & 0.0402 & 84.6\% & 1.35 \\
38 & \texttt{gate\_proj}  & 0.0393 & 73.1\% & 1.66 \\
27 & \texttt{gate\_proj}  & 0.0393 & 73.3\% & 1.65 \\
1  & \texttt{gate\_proj}  & 0.0388 & 85.3\% & 1.33 \\
23 & \texttt{gate\_proj}  & 0.0388 & 81.1\% & 1.45 \\
37 & \texttt{gate\_proj}  & 0.0387 & 77.9\% & 1.53 \\
25 & \texttt{gate\_proj}  & 0.0381 & 80.0\% & 1.47 \\
24 & \texttt{gate\_proj}  & 0.0376 & 80.0\% & 1.47 \\
\bottomrule
\end{tabular}
\caption{The ten largest RL-induced weight changes in the model by Frobenius norm, with each slot's top singular direction's share of that slot's own change energy and its effective rank (maximum possible: 32). Nine of the ten are \texttt{gate\_proj}, consistent with that projection's dominant share of total change energy (Table~\ref{tab:deltaw-module}); all ten are concentrated in 1--2 effective directions.}
\label{tab:svd-top10}
\end{table}

Across all 280 slots, effective rank averages 2.5 (median 2.2, standard deviation 1.1, maximum 8.2). No slot in the model approaches diffuse, full-rank change. The singular vectors themselves, both for these ten slots and for every slot in the model, are retained for a natural causal follow-up: patching a slot's dominant direction into the pre-RL checkpoint and testing whether it recovers part of the post-RL legality gain, or ablating it from the post-RL checkpoint and testing whether legality regresses toward Stage-2's.

\newpage
\bibliographystyle{unsrtnat}
\bibliography{references}

\end{document}